\documentclass[10pt,twocolumn,letterpaper]{article}

\usepackage[pagenumbers]{cvpr} 

\usepackage[dvipsnames]{xcolor}

\usepackage{graphicx}
\usepackage{amsmath}
\usepackage{amssymb}
\usepackage{booktabs}

\usepackage{csvsimple}
\usepackage{amsmath,amssymb} 
\usepackage{xspace}
\usepackage{etoolbox}
\usepackage{makecell}
\usepackage{arydshln}
\usepackage[dvipsnames]{xcolor}

\newif\ifprintcomments
\printcommentstrue
\newcommand{\mysection}[1]{\vspace{3pt}\noindent\textbf{#1.}}

\newcommand{\Table}[1]{Table~\ref{tab:#1}}
\newcommand{\Figure}[1]{Figure~\ref{fig:#1}}
\newcommand{\Equation}[1]{Equation~\eqref{eq:#1}}

\makeatletter
\DeclareRobustCommand\onedot{\futurelet\@let@token\@onedot}
\def\@onedot{\ifx\@let@token.\else.\null\fi\xspace}

\makeatother

\definecolor{cvprblue}{rgb}{0.21,0.49,0.74}
\usepackage[pagebackref,breaklinks,colorlinks,citecolor=cvprblue]{hyperref}

\def\paperID{1430} 
\def\confName{CVPR}
\def\confYear{2024}

\title{SplitNeRF: Split Sum Approximation Neural Field for Joint Geometry, Illumination, and Material Estimation}

\author{Jesus Zarzar \\
KAUST \\
\and
Bernard Ghanem \\
KAUST \\
}

\begin{document}
\maketitle
\begin{abstract}
We present a novel approach for digitizing real-world objects by estimating their geometry, material properties, and environmental lighting from a set of posed images with fixed lighting.
Our method incorporates into Neural Radiance Field (NeRF) pipelines the split sum approximation used with image-based lighting for real-time physical-based rendering.
We propose modeling the scene's lighting with a single scene-specific MLP representing pre-integrated image-based lighting at arbitrary resolutions.
We achieve accurate modeling of pre-integrated lighting by exploiting a novel regularizer based on efficient Monte Carlo sampling.
Additionally, we propose a new method of supervising self-occlusion predictions by exploiting a similar regularizer based on Monte Carlo sampling.
Experimental results demonstrate the efficiency and effectiveness of our approach in estimating scene geometry, material properties, and lighting.
Our method is capable of attaining state-of-the-art relighting quality after only ${\sim}1$ hour of training in a single NVIDIA A100 GPU.

\end{abstract}    
\section{Introduction}
\label{sec:Introduction}

The idea of creating realistic and immersive digital environments has piqued the imagination of countless science fiction authors, science fiction directors, and scientists.
In the past few years, the fields of computer graphics and computer vision have advanced so much that we are capable of creating photo-realistic environments~\cite{carla2017, airsim2017fsr, sim4cv2018}, as well as capturing real-world environments in a way that allows us to render new photo-realistic views~\cite{rematas2022urban, tancik2022blocknerf}.
However, the creation of digital twins~\cite{digital_twins} of objects that can be integrated within photo-realistic environments still requires artists to meticulously hand-design realistic object meshes, materials, and lighting.
While this is feasible for generating a few scenes, large-scale digitization requires automatic ways of reconstructing real-world objects along with their corresponding material properties.

In this work, we address the problem of object inverse rendering: extracting object geometry, material properties, and environment lighting from a set of posed images of the object.
Inverse rendering enables the seamless integration of virtual objects into different environments with varying illumination conditions from simple image captures taken by commonplace camera sensors.

\begin{figure}[t]
 \centering
 \includegraphics[width=1.\linewidth,trim={0 2cm 0 1cm},clip]{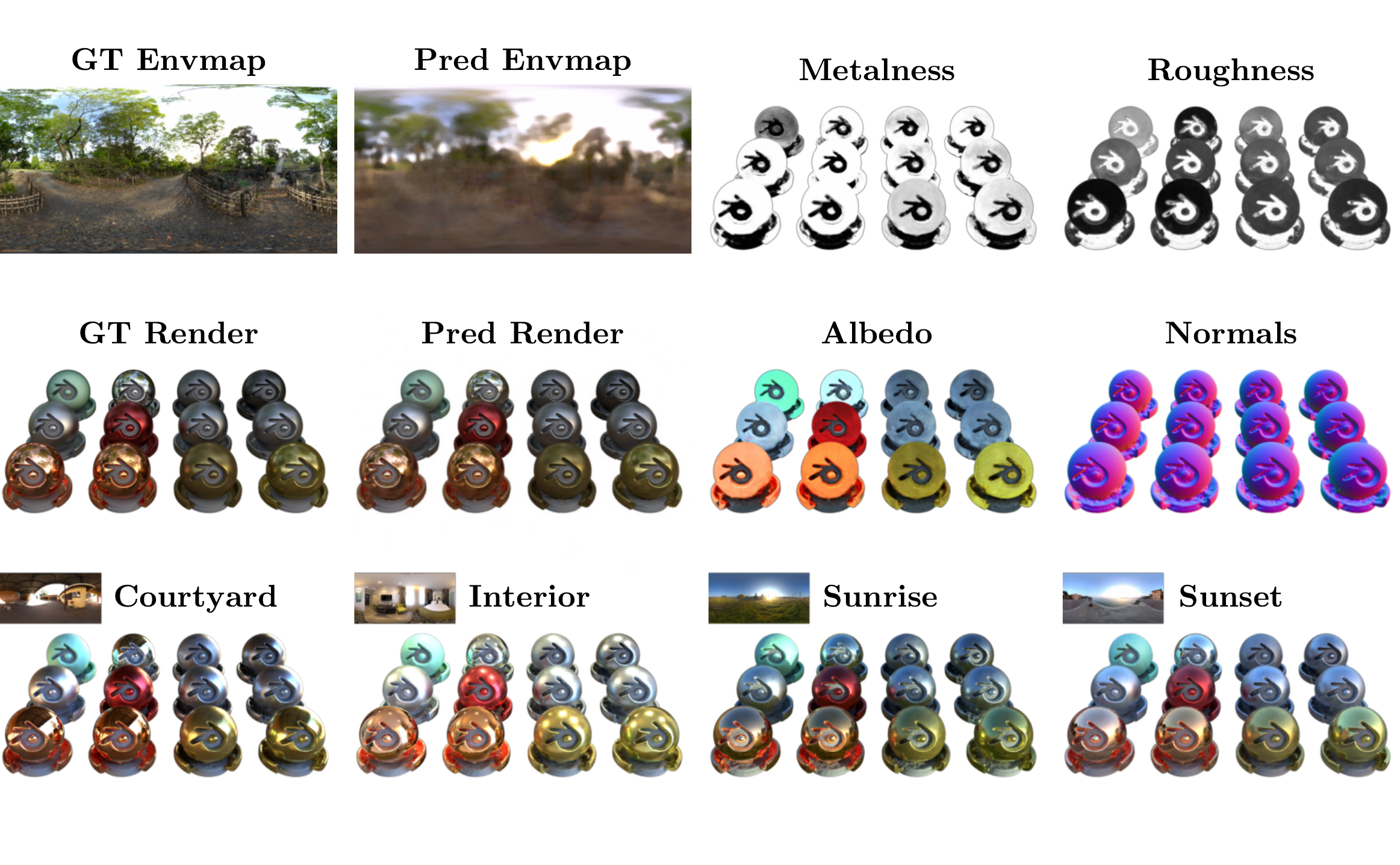}
 \caption{ We visualize the lighting, material properties, and geometry predicted by our model in addition to several relighting predictions of the `materials' scene. Our method is capable of simultaneously predicting high-frequency illumination, material properties (albedo, metalness, and roughness), and geometry.
 }
 \label{fig:pulling}
\end{figure}

Neural rendering methods, such as Neural Radiance Fields (NeRF)~\cite{mildenhall2021nerf, barron2021mipnerf, verbin2022refnerf}, have revolutionized novel view synthesis, 3D reconstruction from images, and inverse rendering.
By directly modeling outgoing radiance at each point in 3D space, NeRF methods excel at accurately recovering scene geometry and synthesizing novel views.
However, a drawback of this approach is that the learned radiance representation entangles environment lighting with the rendered scene's properties, making it challenging to recover material properties and illumination.
Due to the success of NeRFs in reconstructing scenes, several works have proposed modifications to enable inverse rendering~\cite{nerv2021, boss2021neuralpil, mai2023nmf}.
These works build upon NeRF by decomposing radiance into a function of illumination and material properties but differ in their ways of modeling lighting and reflections.
We follow suit with the main goal of efficiency without sacrificing reconstruction quality or the ability to recover high-frequency illumination details.

To achieve these goals, we rely on the split sum approximation~\cite{karis2013real}, which is commonly used in efficient image-based lighting techniques and has been successfully applied for inverse rendering before~\cite{boss2021neuralpil, Munkberg_2022_nvdiffrec}.
This approximation involves splitting the surface reflectance equation into two factors: one responsible for pre-integrating illumination and the other for integrating material properties.
Our first key insight is that this separation allows us to estimate pre-integrated illumination using a Multi-Layer Perceptron (MLP).
This manner of modeling the pre-integrated illumination function is inspired by the modeling of radiance fields, which model a complex integral of lighting and material properties using an MLP.
Correspondingly, our illumination representation inherits beneficial properties observed with the modeling of radiance fields such as smoothness.
To ensure accurate learning of illumination, we introduce a novel regularizer based on Monte Carlo sampling.

However, the split sum approximation on its own does not take into account self-occlusions.
This hinders the estimation of material properties since shadows tend to be incorrectly attributed to being part of an object's albedo.
Thus, we derive an occlusion factor to correctly account for self-occlusions.
This factor is then approximated via Monte Carlo sampling and used to supervise an MLP dedicated to predicting self-occlusions.

Altogether, our method is capable of attaining state-of-the-art relighting results with under an hour of training on a single NVIDIA A100 GPU.

\mysection{Contributions} We claim the following contributions:

\textbf{(i)} We propose a novel representation for representing pre-integrated illumination as a single MLP along with a corresponding regularization to ensure accurate learning.

\textbf{(ii)} We derive a method for approximating the effect of self-occlusions on pre-integrated lighting and use it to supervise an occlusion MLP.

\textbf{(iii)} We demonstrate the effectiveness of our method in extracting environmental lighting and material properties, achieving state-of-the-art relighting quality with under one hour of training on a single NVIDIA A100 GPU.

\section{Related Work}
\label{sec:RelatedWork}

The problem of digitizing real-world objects and environments has long been a subject of active research in computer vision and computer graphics.
We approach this problem through the lenses of neural rendering and neural inverse rendering; paradigms with lots of recent attention.
We now provide a brief overview of related works in these areas.

\subsection{Neural Rendering and 3D Reconstruction}

Novel view synthesis is the task of rendering new views of a scene given a set of observations of the scene.
Neural Radiance Fields (NeRF)~\cite{mildenhall2021nerf} and its variants~\cite{barron2021mipnerf, verbin2022refnerf, muller2022instant, chen2022tensorf, Rojas_2023_ICCV} have demonstrated remarkable success in the task of novel view synthesis.
NeRF directly models the volumetric scene function by predicting radiance and density at each 3D point in space while supervising learning with a photometric reconstruction loss.
Due to its success in implicitly learning accurate 3D reconstructions, several works have branched out to reconstruct accurate meshes through neural rendering~\cite{unisurf2021, sun2022neural}.
Signed Distance Function (SDF)-based methods~\cite{wang2021neus, yariv2021volume, wang2022hf, li2023neuralangelo} model density as a function of the SDF to obtain well-defined surfaces.
By increasing sharpness during training in the conversion from SDF to density these methods can transition from volume rendering to surface rendering as they train.
While effective, these methods suffer from entangled representations of scene geometry, material properties, and lighting.
Our work follows the surface rendering pipeline proposed in~\cite{wang2021neus}, but reformulates the radiance prediction in a manner that disentangles environment lighting and material properties.

\subsection{Neural Inverse Rendering}
The task of inverse rendering consists of estimating the properties of a 3D scene such as shape, material, and lighting from a set of image observations, and is a long-standing problem in computer graphics.
The success of neural rendering methods for novel view rendering and 3D reconstruction has led to a variety of works~\cite{boss2021nerd,zhang2021nerfactor, physg2021, nerv2021, boss2021neuralpil, Munkberg_2022_nvdiffrec, zhang2022invrender, zhang2022iron, liu2023nero, mai2023nmf} exploiting neural rendering for inverse rendering.
Due to the challenging nature of this problem, a wide variety of simplifying assumptions have been adopted.
Some works simplify the modeling of lighting by using low-frequency representations such as spherical gaussians~\cite{boss2021nerd,zhang2021nerfactor, physg2021, nerv2021, zhang2022invrender, NEFII2023, Jin2023TensoIR} or low-resolution environment maps~\cite{boss2021nerd, physg2021, zhang2022invrender}.
While this approximation generally allows for closed-form solutions of the rendering integral, it does not capture natural high-frequency illumination.
Our work leverages the split sum approximation~\cite{karis2013real}, proposed for real-time rendering of image-based global illumination to enable the learning of high-frequency environment lighting.
The split sum approximation has been adopted by several inverse rendering methods~\cite{boss2021neuralpil, Munkberg_2022_nvdiffrec, liu2023nero}.
Pre-integrated lighting has been represented as an autoencoder-based illumination network~\cite{boss2021neuralpil, liang2023envidr}, as a set of learnable images for different roughness levels~\cite{Munkberg_2022_nvdiffrec, neuspir2023}, and as an MLP with integrated spherical harmonic encoding as input~\cite{liu2023nero}.
In contrast, we propose modeling pre-integrated lighting as the output of an MLP paired with a novel regularization, which ensures the network correctly learns to represent pre-integrated lighting.
An issue arising from the split sum approximation is that the pre-integration is blind to geometry and thus does not account for the occlusion of light sources due to geometry at different locations throughout the scene.
Our work tackles this issue by supervising the prediction of ambient occlusion through Monte Carlo sampling.

\section{Methodology}\label{sec:Methodology}

Our method aims to extract a scene's geometry, material properties, and illumination from a set of posed images of the scene.
We accomplish this by incorporating a decomposed formulation of radiance into a surface rendering pipeline.
In the following sections, we begin with an overview of the surface rendering pipeline.
We then detail the physically-based radiance formulation, which allows us to decompose radiance into illumination and material properties.
Next, we describe our proposed MLP representation for illumination along with the additional loss term it requires.
Afterward, we derive a method for estimating an occlusion factor to account for visibility within the split sum approximation.
Finally, we describe additional regularization used to facilitate learning.

\begin{figure}[t]
 \centering
 \includegraphics[width=1.\linewidth]{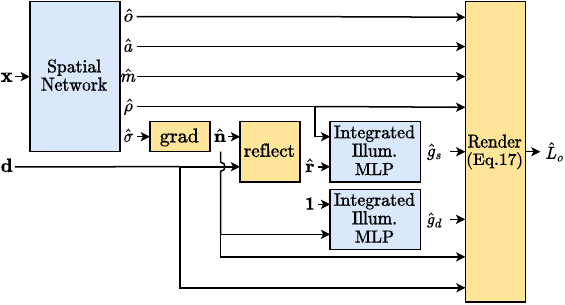}
 \caption{
     \textbf{Proposed Architecture.} A spatial network is used to map spatial coordinates into geometry,  material properties, occlusion, and spatial features. The predicted geometry is used along with predicted roughness by the pre-integrated illumination network to predict both pre-integrated specular and diffuse terms. Finally, the pre-integrated specular and diffuse terms are combined with material properties along with an extra corrective term to produce the output radiance.
 }
 \label{fig:pipeline}
\end{figure}

\subsection{Overview of Neural Rendering}
Neural volume rendering relies on learning two functions: $\sigma(\mathbf{x}; \theta) : \mathbb{R}^3 \mapsto \mathbb{R} $ which maps a point in space $\mathbf{x}$ onto a density $\sigma$, and $\mathbf{L}_o(\mathbf{x}, \mathbf{\omega}_o; \theta) : \mathbb{R}^3 \times \mathbb{R}^3 \mapsto \mathbb{R}^3 $ that maps point $\mathbf{x}$ viewed from direction $\mathbf{\omega}_o$ onto a radiance $\mathbf{L}_o$.
The parameters $\theta$ that define the density and radiance functions are typically optimized to represent a single scene by using multiple posed views of the scene.
To learn these functions, they are evaluated at multiple points along a ray $\mathbf{r}(t) = \mathbf{o} - t \mathbf{\omega}_o$, $t \in [t_n, t_f]$, defined by the camera origin $\mathbf{o} \in \mathbb{R}^3$, pixel viewing direction $\mathbf{\omega}_o$, and camera near and far clipping planes $t_n$ and $t_f$.
A pixel color for the ray can then be obtained through volume rendering via:

\begin{equation}
    \hat{\mathbf{C}}(\mathbf{r}; \theta) = \int\limits_{t_n}^{t_f} T(t) \: \hat{\sigma}(\mathbf{r}(t)) \: \hat{L}_o(\mathbf{r}(t), \mathbf{\omega}_o) \: \mathrm{d}t,
\label{eq:volume_rendering}
\end{equation}
\begin{equation}
    \:\text{where}\:\: T(t) = \exp\left(- \int\limits_{t_n}^t \hat{\sigma}(\mathbf{r}(s)) \: \mathrm{d}s\right).
\label{eq:volume_rendering2}
\end{equation}

In practice, a summation of discrete samples along the ray is used to approximate the integral.
This volume rendering process allows us to supervise the learning of implicit functions $L_o$ and $\sigma$, in a pixel-wise fashion through the reconstruction loss: 
\begin{equation}
    \mathcal{L}_{\text{rec}}(R; \theta) = \frac{1}{|R|} \sum\limits_{\mathbf{r} \in R} \left\| \mathbf{C}(\mathbf{r}) - \hat{\mathbf{C}}(\mathbf{r}; \theta) \right\|^2_2,
\label{eq:loss_rec}
\end{equation}
where $R$ is a batch of rays generated from a random subset of pixels from training images.

The learned geometry can be improved if, instead of directly predicting density $\sigma$, a signed distance field (SDF) is learned and then mapped to density.
To this end, we follow the SDF formulation proposed in NeuS ~\cite{wang2021neus}.
Learning a valid SDF requires the use of an additional Eikonal loss term $\mathcal{L}_{\text{Eik}}$.
For more details, please refer to ~\cite{wang2021neus}.

Since volume density $\sigma$ depends only on a point's position in space while output radiance $L_o$ depends on both position and viewing direction, neural rendering networks are typically split into a spatial network and a radiance network.
As shown in \Figure{pipeline}, we maintain the spatial network to estimate density along with additional material properties but rely on a physically-based~\cite{PBR} radiance estimation instead of a radiance network.

\begin{figure*}[htb]
 \centering
 \includegraphics[width=1.\linewidth]{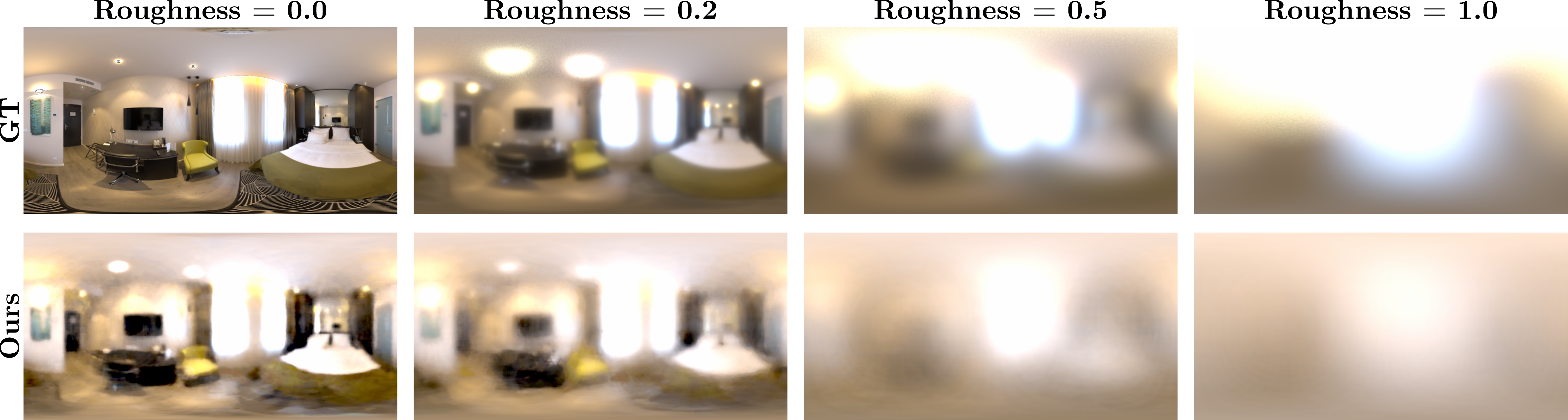}
 \caption{
     \textbf{Pre-Integrated environment illumination.} We visualize the pre-integrated illumination for varying roughness values along with our model's prediction for the `toaster' scene. Our pre-integrated illumination MLP is capable of accurately approximating pre-integrated lighting across roughness values thanks to our novel Monte Carlo regularization loss.
 }
 \label{fig:pre-integrated}
\end{figure*}

\subsection{Physically-Based Rendering}
Given knowledge of a scene's geometry, material properties, and illumination, it is possible to model the outgoing radiance $\mathbf{L}_o(\mathbf{x}, \mathbf{\omega}_o)$ reflected at any position $\mathbf{x}$ of an object's surface in direction $\mathbf{\omega}_o$ by integrating over the hemisphere $\Omega$ defined by the surface's normal $\mathbf{n}$ using the reflectance equation:

\begin{equation}
    \mathbf{L}_o = \int\limits_\Omega (\mathbf{k}_d \frac{\mathbf{a}}{\pi} + \mathbf{f}_s) \mathbf{L}_i \langle \mathbf{\omega}_i,  \mathbf{n} \rangle d\mathbf{\omega}_i ,
\label{eq:light_transport}
\end{equation}

\noindent where $\mathbf{L}_i$ is the incoming radiance, $\mathbf{a}$ is the material's diffuse albedo, and $\mathbf{k}_d$ and $\mathbf{f}_s$ are material properties dependent on the object's Bidirectional Reflectance Distribution Function (BRDF).
For clarity, we omit from the notation the dependency of incoming radiance on $\mathbf{\omega}_i$ as well as the dependency of material properties on position $\mathbf{x}$.
This integral can be split into its diffuse and specular components.

\begin{equation}
    \mathbf{L}_d = \mathbf{k}_d \frac{\mathbf{a}}{\pi} \int\limits_\Omega  \mathbf{L}_i \langle \mathbf{\omega}_i, \mathbf{n} \rangle d \mathbf{\omega}_i , \quad
    \mathbf{L}_s = \int\limits_\Omega \mathbf{f}_s \mathbf{L}_i \langle \mathbf{\omega}_i, \mathbf{n} \rangle d\omega_i .
\label{eq:light_transport_diffuse}
\end{equation}


Computing the specular integral for any general scene is not possible, and approximating it directly using a Monte Carlo simulation is very expensive.
Thus, image-based lighting methods often employ the split sum approximation to calculate specular lighting by splitting the integral into two components: one containing the incoming light $\mathbf{L}_i$, and one that only depends on material properties independent of lighting.
Modeling the BRDF using the Cook-Torrance GGX ~\cite{Torrance, ggx_brdf} model leads to the following approximation for $\mathbf{L}_s$:

\begin{equation}
    \mathbf{L}_s \approx \frac{\int\limits_\Omega D (\mathbf{\omega}_i, \mathbf{\omega}_r, \rho) \mathbf{L}_i \langle \mathbf{\omega}_i,  \mathbf{n} \rangle d\mathbf{\omega}_i}{\int\limits_\Omega D (\mathbf{\omega}_i, \mathbf{\omega}_r, \rho) \langle \mathbf{\omega}_i, \mathbf{n} \rangle d\mathbf{\omega}_i} \int\limits_\Omega \mathbf{f}_s \langle \mathbf{\omega}_i, \mathbf{n} \rangle d\mathbf{\omega}_i ,
\label{eq:split_sum}
\end{equation}

\noindent where $D (\mathbf{\omega}_i, \mathbf{\omega}_r, \rho)$ is the microfacet normal distribution function dependent on the direction of light reflection $\mathbf{\omega}_r$ as well as the surface roughness $\rho$.
The term on the right can be pre-computed since it is independent of a scene's lighting.
We follow the formulation from~\cite{karis2013real} and use a two-dimensional lookup table with precomputed values $F_1$ and $F_2$.
That is, 

\begin{equation}
\begin{split}
     & \int\limits_\Omega \mathbf{f}_s \langle \mathbf{\omega}_i, \mathbf{n} \rangle d\mathbf{\omega}_i = \mathbf{F}_r * F_1 + F_2, \\
     & \mathbf{F}_r = \mathbf{F}_0 + (1 - \rho - \mathbf{F}_0) * (1 - \langle \mathbf{n}, \mathbf{v} \rangle )^5, \\
     & \mathbf{F}_0 = (1 - m) * 0.04 + m * \mathbf{a}, \\
     & \mathbf{k}_d = (1 - m) * (1 - \mathbf{F}_r) , 
\label{eq:diffuse_fresnel}
\end{split}
\end{equation}

\noindent where $m$ and $\rho$ are material properties describing the metalness and roughness of a surface point respectively.
As shown in \Figure{pipeline}, we estimate a material's metalness $\hat{m}$, roughness $\hat{\rho}$, and albedo $\hat{\mathbf{a}}$  as additional outputs from the spatial network.

The term on the left in \Equation{split_sum} depends on the lighting and the chosen microfacet distribution function $D(\mathbf{\omega}_i, \mathbf{\omega}_r, \rho)$, which must be approximated whenever the environment lighting changes.
In the following sections, we refer to this term as $g(\mathbf{\omega}_r, \rho)$.
For a given environment lighting, this term can be pre-integrated and is typically stored in an environmental mipmap where different mipmap levels correspond to varying values of microfacet roughness.

\subsection{MLP Representation}

We propose to estimate the pre-integrated lighting $g(\mathbf{\omega}_r, \rho)$ at different roughness levels through a pre-integrated illumination MLP $\hat{g}(\hat{\mathbf{\omega}}_r, \hat{\rho})$. That is,

\begin{equation}
    \hat{\mathbf{L}}_s = \hat{g}(\hat{\mathbf{\omega}}_r, \hat{\rho}) * (\hat{\mathbf{F}}_r * F_1 + F_2) .
\label{eq:split_sum_nn}
\end{equation}

The pre-integrated lighting $g(\mathbf{\omega}_r, \rho)$  has two special forms for the specific cases of $\rho = 0$ and $\rho = 1$.

\begin{equation}
    g(\mathbf{\omega}, 0) = \mathbf{L}_i(\mathbf{\omega}) , \quad
    g(\mathbf{n}, 1) = \frac{1}{\pi} \int\limits_\Omega \mathbf{L}_i \langle \mathbf{\omega}_i, \mathbf{n} \rangle d\mathbf{\omega}_i,
\label{eq:nn_special}
\end{equation}

This allows us to reuse the network $\hat{g}$ to approximate $\mathbf{L}_d$:
\begin{equation}
    \hat{\mathbf{L}}_d =  \hat{g}(\hat{\mathbf{n}}, 1) \hat{\mathbf{k}}_d \hat{\mathbf{a}},
\label{eq:diffuse_approx}
\end{equation}

The predictions $\hat{g}$ should accurately represent the environment lighting at different levels of roughness.
We achieve this through a loss term based on Monte Carlo estimates $\bar{g}$ of the original integral for varying roughness and reflected directions using the predicted environment map $\hat{\mathbf{L}}_i(\mathbf{\omega}) = \hat{g}(\mathbf{\omega}, 0)$.

\begin{equation}
\begin{split}
    & \mathcal{L}_{\text{D}}(\theta) = \frac{1}{|\mathcal{S}|} \sum\limits_{s \in \mathcal{S}} \left\| \hat{g}(s) - \bar{g}(s) \right\|^2_2, \\
    & \bar{g}(s) = \frac{\sum\limits_{\mathbf{\omega}_i \in \Omega}{D(\mathbf{\omega}_i, \mathbf{\omega}_s, \rho_s) \hat{g}(\mathbf{\omega}_i, 0) \langle \mathbf{\omega}_i, \mathbf{\omega}_s \rangle }}{\sum\limits_{\mathbf{\omega}_i \in \Omega}{D(\mathbf{\omega}_i, \mathbf{\omega}_s, \mathbf{\rho}_s) \langle \mathbf{\omega}_i, \mathbf{\omega}_s \rangle}},
\label{eq:gt_roughness}
\end{split}
\end{equation}

\noindent where the set $\mathcal{S}$ consists of paired samples of directions $\mathbf{\omega}_s$ taken uniformly on a sphere, and roughness samples $\rho_s$ with half the samples taken uniformly in the range $[0, 1]$ and the other half fixed to $1$ to ensure correct learning of diffuse lighting.
The set $\Omega$ of light direction samples is also taken uniformly on a sphere.
While a different sampling could lead to reduced variance, we utilize uniform spherical sampling for $\mathbf{\omega}_i$ to be more computationally efficient.
Uniform spherical sampling allows us to share light samples across the batch of predictions, thus reducing the number of evaluation calls to the light function $\hat{g}(\mathbf{\omega}, 0)$.
We visualize both $g$ and $\hat{g}$ in \Figure{pre-integrated} for a specific scene.

\subsection{Occlusion Factors}
The split sum approximation does not consider the occlusion of light sources due to geometry.
To incorporate occlusions, incoming light $\mathbf{L}_i$ would need to be multiplied by a binary visibility function $V_i$ as follows: 
\begin{equation}
    \mathbf{L}^V_d = \mathbf{k}_d \frac{\mathbf{a}}{\pi} \int\limits_\Omega  \mathbf{L}_i V_i \langle \mathbf{\omega}_i, \mathbf{n} \rangle d\mathbf{\omega}_i ,
\label{eq:diffuse_visibility}
\end{equation}

\noindent with $V_i$ taking a value of $1$ when there are no occlusions and $0$ when incoming light is occluded by geometry.
The integral can be written as an occlusion factor $\mathbf{o}_d(\mathbf{x})$ multiplying the split sum diffuse light term from \Equation{light_transport_diffuse}:

\begin{equation}
\begin{split}
    & \int\limits_\Omega  \mathbf{L}_i V_i \langle \mathbf{\omega_i}, \mathbf{n} \rangle d\mathbf{\omega}_i = \frac{\int\limits_\Omega  \mathbf{L}_i V_i \langle \mathbf{\omega}_i, \mathbf{n} \rangle d\mathbf{\omega}_i }{\int\limits_\Omega  \mathbf{L}_i \langle \mathbf{\omega}_i, \mathbf{n} \rangle d\mathbf{\omega}_i } \int\limits_\Omega  \mathbf{L}_i \langle \mathbf{\omega}_i, \mathbf{n} \rangle d\mathbf{\omega}_i , \\
    & \mathbf{L}^V_d = \mathbf{o}_d(\mathbf{x}) \mathbf{L}_d , \quad
    \mathbf{o}_d(\mathbf{x}) = \frac{\int\limits_\Omega  \mathbf{L}_i V_i \langle \mathbf{\omega}_i, \mathbf{n} \rangle d\mathbf{\omega}_i }{\int\limits_\Omega  \mathbf{L}_i \langle \mathbf{\omega}_i, \mathbf{n} \rangle d\mathbf{\omega}_i }
\label{eq:diffuse_visibility_factor}
\end{split}
\end{equation}

We propose learning the occlusion factor $\mathbf{o}_d(\mathbf{x})$ with an MLP.
The learnt occlusion term $\hat{\mathbf{o}}_d(\mathbf{x})$ is then supervised by Monte Carlo estimates $\bar{\mathbf{o}}_d(\mathbf{x})$ using the predicted geometry.

\begin{equation}
    \bar{\mathbf{o}}_d(\mathbf{x}) = \frac{\sum\limits_{\mathbf{\omega}_i \in \Omega} \mathbf{L}_i V_i}{\sum\limits_{\mathbf{\omega}_i \in \Omega} \mathbf{L}_i},
\label{eq:diffuse_visibility_approx}
\end{equation}

\noindent with $\mathbf{\omega}_i$ taken from a cos-weighted sampling of the hemisphere around the normal at $\mathbf{x}$.
A similar derivation can be followed for the specular occlusion term leading to the following Monte Carlo estimate $\bar{\mathbf{o}}_s(\mathbf{x})$:

\begin{equation}
    \bar{\mathbf{o}}_s(\mathbf{x}) = \frac{\sum\limits_{\mathbf{\omega}_i \in \Omega} \mathbf{L}_i V_i \langle \mathbf{\omega}_i, \mathbf{n} \rangle }{\sum\limits_{\mathbf{\omega}_i \in \Omega} \mathbf{L}_i \langle \mathbf{\omega}_i, \mathbf{n} \rangle},
\label{eq:specular_visibility}
\end{equation}

\noindent with $\mathbf{\omega}_i$ sampled from the GGX distribution centered around the normal at $\mathbf{x}$.
Given the Monte Carlo estimates $\bar{\mathbf{o}}_d(\mathbf{x})$ and $\bar{\mathbf{o}}_s(\mathbf{x})$, we supervise the predicted occlusion terms $\hat{\mathbf{o}}_d(\mathbf{x})$ and $\hat{\mathbf{o}}_s(\mathbf{x})$ as follows:

\begin{equation}
    \mathcal{L}_{\text{o}}(\theta) = \frac{1}{|\mathcal{X}|} \sum\limits_{x \in \mathcal{X}} w \left\| \hat{\mathbf{o}}(\mathbf{x}) - \bar{\mathbf{o}}(\mathbf{x}) \right\|^2_2,
\label{eq:loss_occlusion}
\end{equation}

\noindent where the sample set $\mathcal{X}$ is a random subset of the points sampled for volume rendering, and the weights $w$ are the corresponding normalized volume rendering weights.
Weighting the loss function by the volume rendering weights is required so that the occlusion prediction focuses only on learning surface points.

The output radiance at each point in space is thus calculated as follows:
\begin{equation}
    \hat{\mathbf{L}}_o =  \gamma( \hat{\mathbf{o}}_d * \hat{\mathbf{L}}_d + \hat{\mathbf{o}}_s * \hat{\mathbf{L}}_s ),
\label{eq:output_radiance_approx}
\end{equation}

\noindent where $\gamma$ is a function mapping the predicted output radiance $\hat{\mathbf{L}}_o$ from linear to SRGB space.

\subsection{Material Regularization}
To better learn material properties, we introduce a soft regularizer to reduce the prediction of metallic materials.
This encourages the model to prefer explaining outgoing radiance through albedo and roughness whilst still allowing the prediction of metallic materials.
We implement this regularization as a weighted $L_2$ loss with the same weighting as for the occlusion loss in \Equation{loss_occlusion}.
That is,

\begin{equation}
    \mathcal{L}_{\text{m}}(\theta) = \frac{1}{|\mathcal{X}|} \sum\limits_{\mathbf{x} \in \mathcal{X}} w \left\|  \hat{m}(\mathbf{x}) \right|^2_2.
\label{eq:loss_metallic}
\end{equation}

\begin{table*}[htb]
    \resizebox{\textwidth}{!}{
        \begin{filecontents*}{results/relighting_nerfactor.csv}
,Name,drumsLPIPS,ficusLPIPS,hotdogLPIPS,legoLPIPS,avgLPIPS,drumsSSIM,ficusSSIM,hotdogSSIM,legoSSIM,avgSSIM,drumsPSNR,ficusPSNR,hotdogPSNR,legoPSNR,avgPSNR
0,NerFactor,0.130,0.090,0.118,0.141,0.120,0.879,0.932,0.914,0.854,0.895,20.01,23.71,26.15,24.77,23.66
1,NVDiffRec,0.097,0.085,0.124,0.137,0.111,0.890,0.907,0.892,0.831,0.880,20.72,20.09,24.64,22.09,21.88
2,NVDiffRecMC,0.094,0.079,0.089,0.134,0.099,0.899,0.910,0.938,0.862,0.902,21.56,21.38,\textbf{29.05},24.24,24.06
3,NMF,0.075,0.063,0.120,0.116,0.093,0.906,0.934,0.876,0.863,0.895,21.54,21.36,22.47,23.58,22.23
4,NERO,0.110,0.062,0.093,0.108,0.093,0.900,0.936,0.908,0.884,0.907,20.73,23.58,25.28,25.14,23.68
5,Ours,\textbf{0.058},\textbf{0.041},\textbf{0.069},\textbf{0.075},\textbf{0.061},\textbf{0.935},\textbf{0.964},\textbf{0.947},\textbf{0.920},\textbf{0.941},\textbf{24.72},\textbf{27.45},29.04,\textbf{28.02},\textbf{27.31}
\end{filecontents*}
\csvreader[
            tabular=l|c|cccc|c|cccc|c|cccc,
            table head={
                \Xhline{2\arrayrulewidth}
                & \multicolumn{5}{c|}{$\mathbf{PSNR} \uparrow$} & \multicolumn{5}{c|}{$\mathbf{SSIM} \uparrow$} & \multicolumn{5}{c}{$\mathbf{LPIPS} \downarrow$} \\
                \Xhline{2\arrayrulewidth}
                \bfseries & \bfseries avg. & \bfseries drums & \bfseries ficus & \bfseries hotdog & \bfseries lego & \bfseries avg. & \bfseries drums & \bfseries ficus & \bfseries hotdog & \bfseries lego & \bfseries avg. & \bfseries drums & \bfseries ficus & \bfseries hotdog & \bfseries lego
                \\\Xhline{2\arrayrulewidth}
            },
            table foot=\Xhline{2\arrayrulewidth},
            head to column names,
        ]{results/relighting_nerfactor.csv}{}
        {\bfseries\Name & \avgPSNR & \drumsPSNR & \ficusPSNR & \hotdogPSNR & \legoPSNR & \avgSSIM & \drumsSSIM & \ficusSSIM & \hotdogSSIM & \legoSSIM & \avgLPIPS & \drumsLPIPS & \ficusLPIPS & \hotdogLPIPS & \legoLPIPS}
    }
    \caption{\textbf{NeRFactor Relighting Metrics.} We evaluate the relighting quality of our method against the baselines using $20$ test images and $8$ low-frequency illumination maps from the NeRFactor dataset. Images are scaled by a per-channel factor before computing metrics. Our method outperforms the baselines across all reconstruction metrics for all but one scene.}
    \label{tab:relighting_quantitative}
\end{table*}

\begin{table}[htb]
    \resizebox{\linewidth}{!}{
        \begin{filecontents*}{results/relighting_nerf.csv}
,Name,chairLPIPSNerf,drumsLPIPSNerf,ficusLPIPSNerf,hotdogLPIPSNerf,legoLPIPSNerf,materialsLPIPSNerf,micLPIPSNerf,shipLPIPSNerf,avgLPIPSNerf,chairSSIMNerf,drumsSSIMNerf,ficusSSIMNerf,hotdogSSIMNerf,legoSSIMNerf,materialsSSIMNerf,micSSIMNerf,shipSSIMNerf,avgSSIMNerf,chairPSNRNerf,drumsPSNRNerf,ficusPSNRNerf,hotdogPSNRNerf,legoPSNRNerf,materialsPSNRNerf,micPSNRNerf,shipPSNRNerf,avgPSNRNerf,carLPIPSShiny,coffeeLPIPSShiny,helmetLPIPSShiny,teapotLPIPSShiny,toasterLPIPSShiny,avgLPIPSShiny,carSSIMShiny,coffeeSSIMShiny,helmetSSIMShiny,teapotSSIMShiny,toasterSSIMShiny,avgSSIMShiny,carPSNRShiny,coffeePSNRShiny,helmetPSNRShiny,teapotPSNRShiny,toasterPSNRShiny,avgPSNRShiny
0,NVDiffRec,0.099,0.134,0.083,0.141,0.141,0.132,0.092,0.283,0.138,0.886,0.852,0.902,0.894,0.834,0.866,0.927,0.695,0.857,21.70,19.41,20.32,21.17,21.84,21.26,18.48,16.69,20.11,0.102,0.237,0.209,0.046,0.290,0.177,0.913,0.799,0.848,0.972,0.705,0.848,24.85,18.29,19.07,29.39,15.36,21.39
1,NVDiffRecMC,0.084,0.135,0.093,0.118,0.147,0.112,0.097,0.306,0.136,0.918,0.877,0.902,\textbf{0.930},0.865,0.904,0.924,0.750,0.884,24.64,20.57,21.27,\textbf{26.37},\textbf{24.77},24.57,\textbf{18.95},\textbf{18.91},22.50,0.101,\textbf{0.195},0.182,0.039,0.241,0.151,0.917,\textbf{0.921},0.908,0.983,0.827,\textbf{0.911},24.25,\textbf{22.93},22.86,32.70,20.25,24.60
2,NMF,0.093,0.103,0.066,0.135,0.108,0.081,0.087,0.271,0.118,0.908,0.890,0.934,0.890,0.863,0.913,0.926,0.722,0.881,22.53,21.38,22.62,20.52,22.05,24.87,18.34,17.41,21.21,0.092,0.208,\textbf{0.142},0.032,\textbf{0.206},\textbf{0.136},0.917,0.843,\textbf{0.946},0.982,\textbf{0.849},0.908,24.58,17.88,\textbf{27.48},30.66,\textbf{20.41},24.20
3,Ours,\textbf{0.065},\textbf{0.096},\textbf{0.050},\textbf{0.116},\textbf{0.097},\textbf{0.075},\textbf{0.077},\textbf{0.269},\textbf{0.106},\textbf{0.937},\textbf{0.908},\textbf{0.952},0.914,\textbf{0.901},\textbf{0.930},\textbf{0.937},\textbf{0.769},\textbf{0.906},\textbf{25.00},\textbf{22.62},\textbf{26.40},20.94,23.88,\textbf{25.38},18.75,18.88,\textbf{22.73},\textbf{0.072},0.204,0.195,\textbf{0.017},0.234,0.144,\textbf{0.942},0.883,0.877,\textbf{0.993},0.827,0.904,\textbf{26.86},18.70,21.51,\textbf{38.13},19.62,\textbf{24.96}
\end{filecontents*}
\csvreader[
            tabular=l|ccc|ccc,
            table head={
                \Xhline{2\arrayrulewidth}
                & \multicolumn{3}{c|}{$\mathbf{Blender}$} & \multicolumn{3}{c}{$\mathbf{Shiny ~ Blender}$} \\
                \Xhline{2\arrayrulewidth}
                \bfseries & \bfseries PSNR $\uparrow$ & \bfseries SSIM $\uparrow$ & \bfseries LPIPS $\downarrow$ & \bfseries PSNR $\uparrow$ & \bfseries SSIM $\uparrow$ & \bfseries LPIPS $\downarrow$
                \\\Xhline{2\arrayrulewidth}
            },
            table foot=\Xhline{2\arrayrulewidth},
            head to column names,
        ]{results/relighting_nerf.csv}{}
        {\bfseries\Name & \avgPSNRNerf & \avgSSIMNerf & \avgLPIPSNerf & \avgPSNRShiny & \avgSSIMShiny & \avgLPIPSShiny}
    }
    \caption{\textbf{Blender and Shiny Blender Relighting Metrics.}  We report the average relighting reconstruction metrics across all scenes for our extended Blender and Shiny Blender datasets. Metrics are computed as the average of $20$ test views across $7$ high-frequency illumination conditions. Images are scaled by a per-channel factor before computing metrics. Our method outperforms the baselines across all metrics for the Blender dataset and has a higher PSNR for the Shiny Blender dataset. }
    \label{tab:relighting_nerf}
\end{table}

\begin{figure}[htb]
 \centering
 \includegraphics[width=1.\linewidth,trim={0.5cm 2cm 4cm 0},clip]{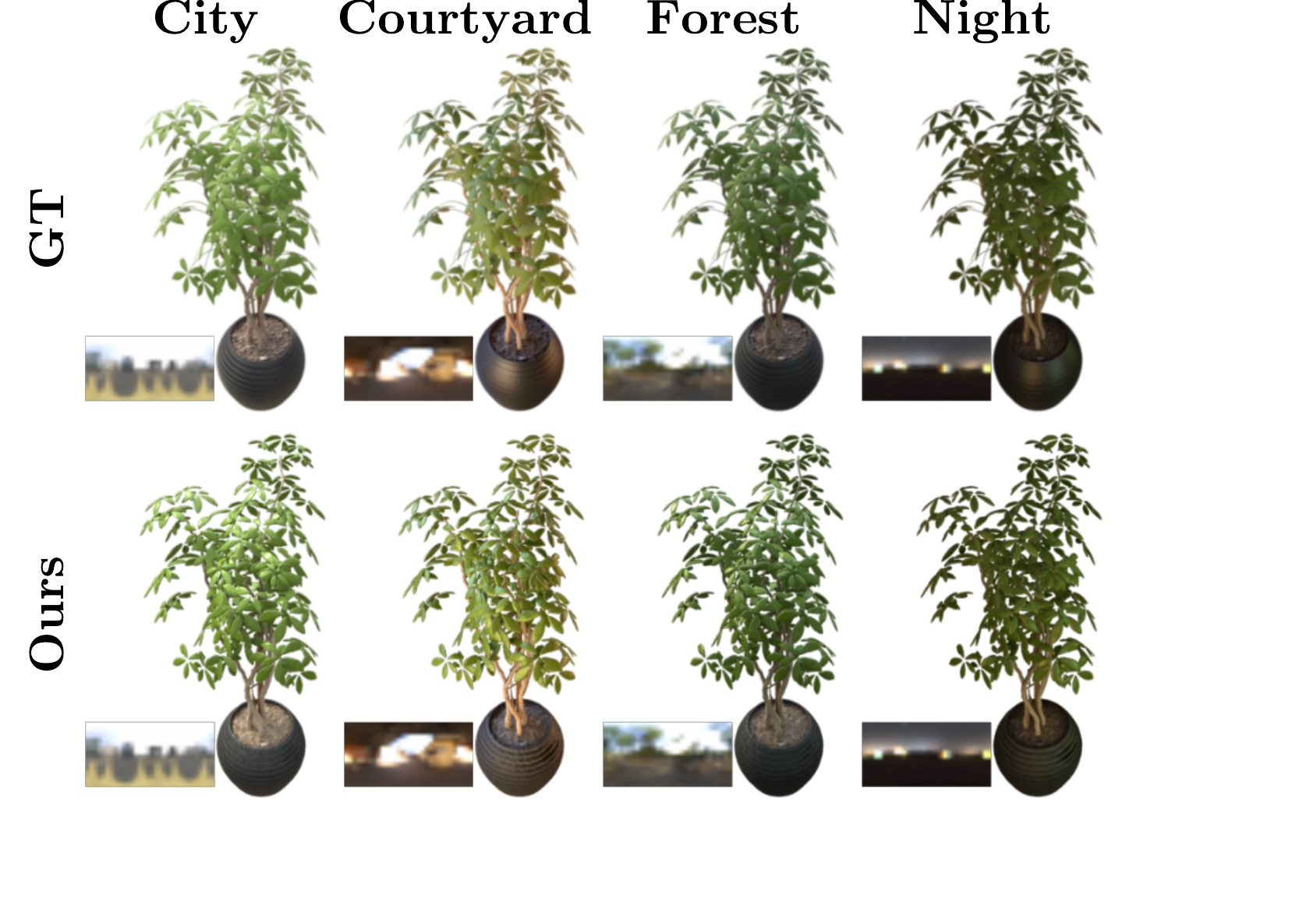}
 \caption{
     \textbf{Qualitative Relighting Results.} We render the predicted mesh and material properties from the `ficus' scene using Blender. Four different low-frequency environment maps from the NeRFactor dataset are visualized.
 }
 \label{fig:relighting_qualitative}
\end{figure}

\begin{table*}[htb]
    \resizebox{\textwidth}{!}{
        \begin{filecontents*}{results/albedo_nerfactor.csv}
,Name,drumsLPIPS,ficusLPIPS,hotdogLPIPS,legoLPIPS,avgLPIPS,drumsSSIM,ficusSSIM,hotdogSSIM,legoSSIM,avgSSIM,drumsPSNR,ficusPSNR,hotdogPSNR,legoPSNR,avgPSNR
0,NerFactor,0.132,0.098,\textbf{0.093},\textbf{0.112},0.109,0.878,0.923,0.937,\textbf{0.903},0.910,20.75,22.05,27.75,\textbf{23.58},23.53
1,NVDiffRec,0.120,0.099,0.108,0.172,0.125,0.880,0.924,0.944,0.842,0.898,19.66,22.64,28.35,21.20,22.96
2,NVDiffRecMC,0.116,0.106,0.101,0.136,0.115,0.895,0.921,\textbf{0.950},0.903,0.917,20.40,22.14,\textbf{29.61},23.17,23.83
3,NeRO,0.138,0.125,0.125,0.176,0.141,0.880,0.900,0.883,0.821,0.871,20.00,20.42,23.27,21.01,21.18
4,Ours,\textbf{0.099},\textbf{0.051},0.121,0.160,\textbf{0.108},\textbf{0.922},\textbf{0.974},0.929,0.870,\textbf{0.924},\textbf{23.73},\textbf{29.41},24.68,23.33,\textbf{25.29}
\end{filecontents*}
\csvreader[
            tabular=l|c|cccc|c|cccc|c|cccc,
            table head={
                \Xhline{2\arrayrulewidth}
                & \multicolumn{5}{c|}{$\mathbf{PSNR} \uparrow$} & \multicolumn{5}{c|}{$\mathbf{SSIM} \uparrow$} & \multicolumn{5}{c}{$\mathbf{LPIPS} \downarrow$} \\
                \Xhline{2\arrayrulewidth}
                \bfseries & \bfseries avg. & \bfseries drums & \bfseries ficus & \bfseries hotdog & \bfseries lego & \bfseries avg. & \bfseries drums & \bfseries ficus & \bfseries hotdog & \bfseries lego & \bfseries avg. & \bfseries drums & \bfseries ficus & \bfseries hotdog & \bfseries lego
                \\\Xhline{2\arrayrulewidth}
            },
            table foot=\Xhline{2\arrayrulewidth},
            head to column names,
        ]{results/albedo_nerfactor.csv}{}
        {\bfseries\Name & \avgPSNR & \drumsPSNR & \ficusPSNR & \hotdogPSNR & \legoPSNR & \avgSSIM & \drumsSSIM & \ficusSSIM & \hotdogSSIM & \legoSSIM & \avgLPIPS & \drumsLPIPS & \ficusLPIPS & \hotdogLPIPS & \legoLPIPS}
    }
    \caption{\textbf{Quantitative Albedo Metrics.} We report the albedo reconstruction quality of our method compared to the baselines using the NeRFactor dataset. Albedo is scaled by a per-channel factor to minimize error. On average, we outperform all baselines across all metrics.}
    \label{tab:albedo_quantitative}
\end{table*}
\section{Experiments}
\label{sec:Experiments}

\subsection{Baselines}
We compare against Nerfactor~\cite{zhang2021nerfactor}, NVDiffRec~\cite{Munkberg_2022_nvdiffrec}, NVDiffRecMC~\cite{hasselgren2022nvdiffrecmc}, NMF~\cite{mai2023nmf}, and NeRO~\cite{liu2023nero}.
Due to the differing evaluation methodologies among these works, we train all baseline methods following publicly released code and report metrics as detailed in the following sections.

\subsection{Experimental setup}
\mysection{Datasets}
We report results using the NeRFactor ~\cite{zhang2021nerfactor} dataset along with extended versions of the NeRF Blender~\cite{mildenhall2021nerf} (Blender) and the RefNeRF Shiny Blender~\cite{verbin2022refnerf} (Shiny Blender) datasets.
The NeRFactor dataset consists of four synthetic scenes, where test images are rendered under eight different low-frequency lighting conditions.
The Blender dataset consists of eight synthetic scenes representing a mix of glossy, specular, and Lambertian objects, while the Shiny Blender dataset consists of six highly reflective synthetic scenes.
To showcase the ability of our model to estimate high-frequency environment lighting, we extend the Blender and Shiny Blender datasets by rendering all objects under seven novel high-frequency lighting conditions.
All models are trained using $100$ posed images, and evaluated on $20$ test images consisting of novel views for each lighting condition.

\mysection{Implementation Details}
We utilize an efficient implementation of NeuS~\cite{instant-nsr-pl} as our surface rendering pipeline.
We train our models for $20,000$ steps using a warmup learning rate scheduler for the first $500$ steps followed by an exponential decay scheduler.
After every $2000$ steps, we estimate the current geometry by using marching cubes~\cite{lorensen1987marching} to extract the isosurface at SDF level-set $0$.
The estimated geometry is used with $64$ samples for the Monte Carlo estimation of occlusion factors.
We use a random subset of $10\%$ of the points from volume rendering to supervise the occlusion network to reduce time and memory requirements.
$8129$ light samples are used for computing illumination loss Monte Carlo estimates.
The final loss is calculated as a linear combination of the proposed losses, with the following coefficients: $\lambda_{\text{rec}} = 10.0$, $\lambda_{D} = 10.0$, $\lambda_{o} = 0.01$, $\lambda_{\text{Eik}} = 0.1$, and $\lambda_{m} = 0.001$.
We run all experiments on a single A100 GPU for a total training time of ${\sim}1$ hour.

\begin{figure*}[htb]
 \centering
 \includegraphics[width=1.\linewidth]{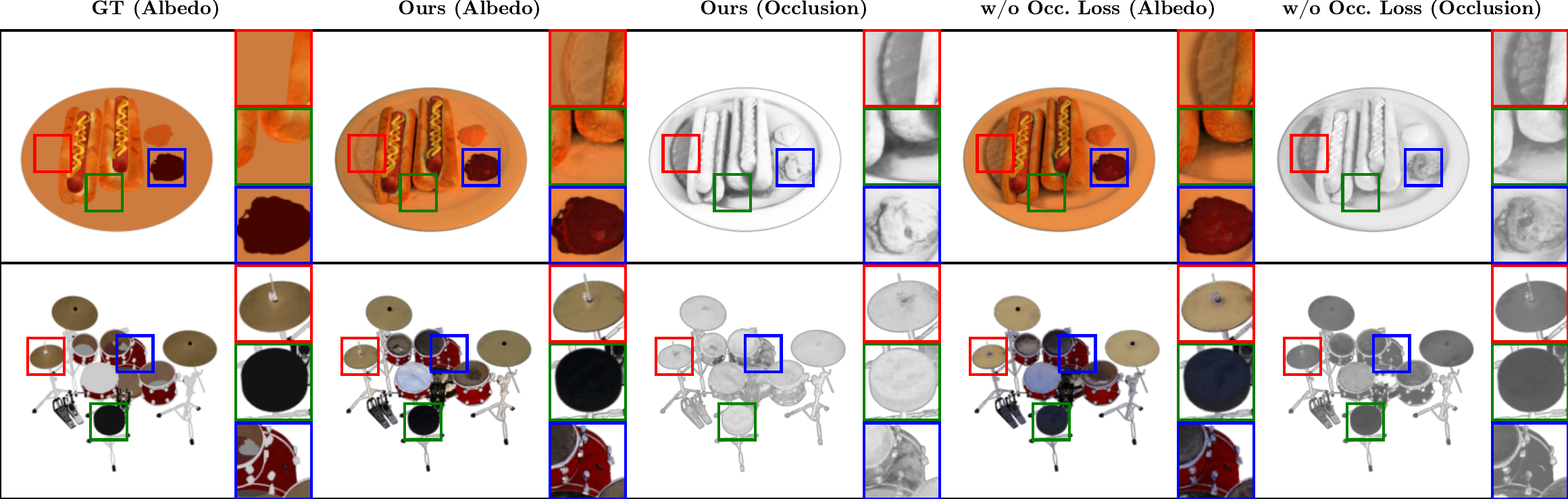}
 \caption{
     \textbf{Occlusion Loss Visualization.} We visualize the albedo and occlusion predicted by our method with and without the proposed occlusion regularization loss. When no regularization is used, we observe that the occlusion prediction fails at disentangling shadows from the albedo. Additionally, darker materials might wind up with lighter albedos due to occlusion overcompensation.
 }
 \label{fig:qualitative_ao}
\end{figure*}

\begin{table}[htb]
    \resizebox{\linewidth}{!}{
        \begin{filecontents*}{results/ablation_nerfactor.csv}
,Unnamed: 0,Name,drumsLPIPSAlbedo,ficusLPIPSAlbedo,hotdogLPIPSAlbedo,legoLPIPSAlbedo,avgLPIPSAlbedo,drumsSSIMAlbedo,ficusSSIMAlbedo,hotdogSSIMAlbedo,legoSSIMAlbedo,avgSSIMAlbedo,drumsPSNRAlbedo,ficusPSNRAlbedo,hotdogPSNRAlbedo,legoPSNRAlbedo,avgPSNRAlbedo,Name.1,drumsLPIPS,ficusLPIPS,hotdogLPIPS,legoLPIPS,avgLPIPS,drumsSSIM,ficusSSIM,hotdogSSIM,legoSSIM,avgSSIM,drumsPSNR,ficusPSNR,hotdogPSNR,legoPSNR,avgPSNR
0,0,Ours,\textbf{0.099},\textbf{0.051},0.121,0.160,\textbf{0.108},\textbf{0.922},\textbf{0.974},0.929,0.870,\textbf{0.924},\textbf{23.73},\textbf{29.41},24.68,23.33,25.29,Ours,0.058,0.041,0.069,0.075,0.061,0.935,0.964,0.947,0.920,0.941,24.72,27.45,29.04,28.02,27.31
1,1,Ours (w/o Occ. Avg.),0.104,0.058,0.116,\textbf{0.159},0.109,0.918,0.963,0.926,0.869,0.919,23.45,27.73,24.24,23.51,24.73,Ours (w/o Occ. Avg.),0.058,\textbf{0.040},0.069,0.075,0.061,0.936,0.964,0.946,0.919,0.941,24.75,27.36,28.93,27.94,27.25
2,2,Ours (w/o Occ. Loss),0.112,0.091,0.125,0.162,0.122,0.907,0.928,0.920,0.844,0.900,21.38,22.86,23.10,21.20,22.13,Ours (w/o Occ. Loss),0.061,0.044,0.068,0.082,0.064,0.935,0.960,0.946,0.903,0.936,24.83,25.82,29.00,25.97,26.40
3,3,Ours (w/o Met. Reg.),0.099,0.057,0.125,0.210,0.123,0.918,0.965,0.918,0.864,0.916,22.62,28.74,22.87,21.47,23.92,Ours (w/o Met. Reg.),0.058,0.042,0.078,0.086,0.066,0.934,0.962,0.938,0.909,0.936,24.58,27.16,27.53,26.96,26.56
4,4,Ours (Mipmap),0.103,0.067,\textbf{0.106},0.180,0.114,0.922,0.955,\textbf{0.940},\textbf{0.881},0.924,23.48,26.70,\textbf{26.91},\textbf{24.61},\textbf{25.43},Ours (Mipmap),\textbf{0.057},0.040,\textbf{0.062},\textbf{0.074},\textbf{0.058},\textbf{0.938},\textbf{0.965},\textbf{0.951},\textbf{0.924},\textbf{0.944},\textbf{24.98},\textbf{27.48},\textbf{30.08},\textbf{28.70},\textbf{27.81}
\end{filecontents*}
\csvreader[
            tabular=l|ccc|ccc,
            table head={
                \Xhline{2\arrayrulewidth}
                & \multicolumn{3}{c|}{$\mathbf{Relighting}$} & \multicolumn{3}{c}{$\mathbf{Albedo}$} \\
                \Xhline{2\arrayrulewidth}
                
                \bfseries & \bfseries PSNR $\uparrow$ & \bfseries SSIM $\uparrow$ & \bfseries LPIPS $\downarrow$ & \bfseries PSNR $\uparrow$ & \bfseries SSIM $\uparrow$ & \bfseries LPIPS $\downarrow$
                \\\Xhline{2\arrayrulewidth}
            },
            table foot=\Xhline{2\arrayrulewidth},
            head to column names,
        ]{results/ablation_nerfactor.csv}{}
        {\bfseries\Name & \avgPSNR & \avgSSIM & \avgLPIPS & \avgPSNRAlbedo & \avgSSIMAlbedo & \avgLPIPSAlbedo }
    }
    \caption{\textbf{NeRFactor Ablation Results.} We report relighting and ablation metrics for different variations of our methodology on the NeRFactor dataset. While employing a mipmap representation for illumination provides improved quality for both relighting and albedo, it comes at a cost of ${\sim}36\%$ higher training time. All other proposed components of our method improve quality.}
    \label{tab:ablation_nerfactor}
\end{table}

\subsection{Relighting}
We extract geometry from our model in the form of a triangular mesh by using marching cubes~\cite{lorensen1987marching}.
At each predicted mesh vertex, we estimate material properties in the form of an albedo, metalness, and roughness.
We then render the predicted geometry using Blender's~\cite{blender2018} physically based shader.
Material properties across faces are obtained by interpolating the predicted vertex material properties.
For baselines where explicit meshes and material properties are extracted, we utilize the same Blender rendering pipeline to compute relighting metrics.
Otherwise, predictions are rendered using the provided relighting methodology.
Before evaluating metrics, a per-channel scaling factor is computed for each scene to compensate for the albedo-lighting ambiguity.
We evaluate the predicted scenes for the NeRFactor, Blender, and Shiny Blender datasets and report the average Peak
Signal-to-Noise Ratio (PSNR), Structural Similarity Index
Measure (SSIM)~\cite{SSIM}, and Learned Perceptual Image Patch
Similarity (LPIPS)~\cite{LPIPS} in \Table{relighting_quantitative} and \Table{relighting_nerf}.
Metrics are reported as an average across $20$ test images and across all illumination maps for each dataset.
This metric gives an aggregated performance measure for geometry and material property estimation.
While it does not measure performance for estimating illumination, an accurate illumination estimation is essential for recovering material properties.
It can be observed that our method outperforms all baselines by a significant margin.
Additionally, we provide renderings of our method's predictions for the NeRFactor dataset's ficus scene under four different illumination conditions in \Figure{relighting_qualitative}.


\subsection{Albedo}
In addition to overall relighting quality, we evaluate the ability of our method to recover albedo.
We report reconstruction metrics on the predicted albedo in \Table{albedo_quantitative}.
As with the relighting metrics, we apply a per-scaling factor to the albedo predictions before computing reconstruction metrics.
Metrics are reported as an average across all $20$ test images for each scene in the NeRFactor dataset.
We exclude results from NMF~\cite{mai2023nmf} since the albedo in their lighting formulation is not comparable to that of the other methods.
Thanks to our proposed occlusion factor, our method is on average better able to reconstruct albedo.



\begin{figure*}[htb]
 \centering
 \includegraphics[width=1.\linewidth]{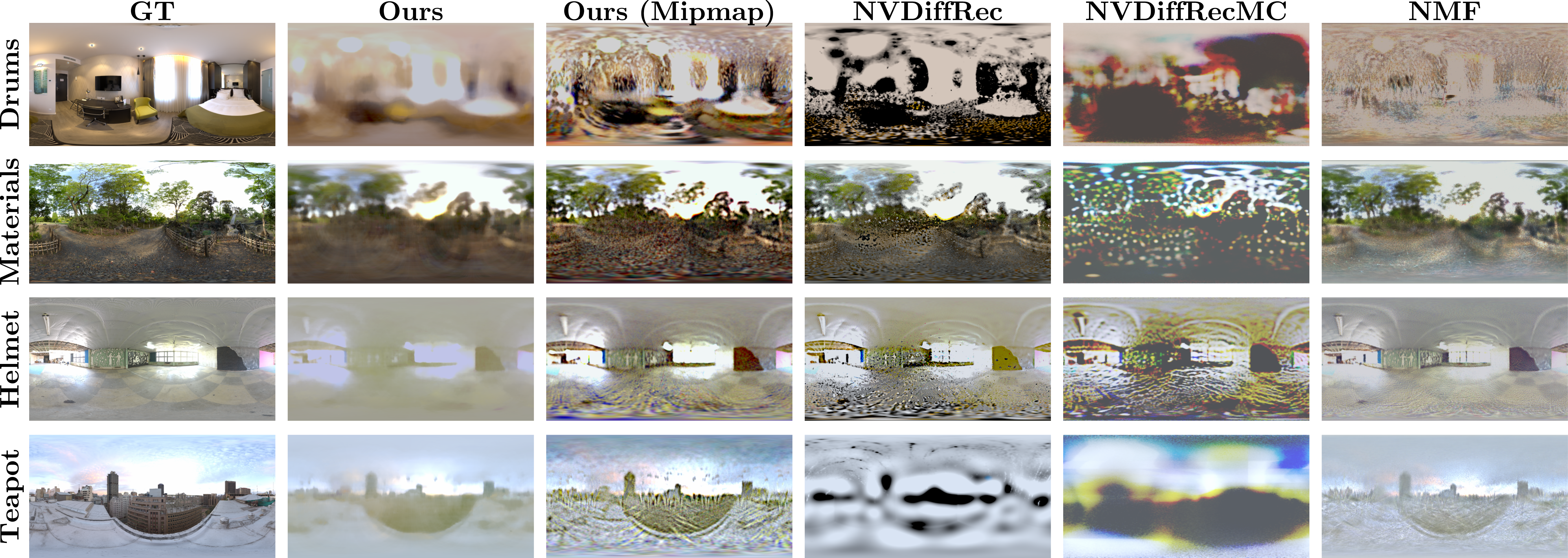}
 \caption{
     \textbf{Blender and Shiny Blender Illumination Visualizations.} We visualize the predicted illumination for our method, our method using mipmap illumination, and baselines for two scenes in the Blender dataset and two scenes in the Shiny Blender dataset. Illumination is scaled by a per-channel factor. Our proposed illumination inherits smoothness from the MLP representation but is still able to capture high-quality details such as trees and buildings.
 }
 \label{fig:envmaps}
\end{figure*}

\section{Discussions}
\label{sec:Discussion}

\subsection{Ablations}

\mysection{Occlusion loss}
We visualize the effects of the proposed occlusion loss in \Figure{qualitative_ao}.
Learning an occlusion factor without supervision leads to errors in the albedo predictions due to the inability to disentangle shadows from object color.
By explicitly supervising an occlusion factor we observe better albedo color predictions such as visualized in the blue box in the hotdog example, and all boxes in the drums example.
Additionally, shadows are better disentangled from albedo as observed in the red and green boxes for the hotdog example.
Quantitatively, we measure the importance of adding the occlusion loss to our model in \Table{ablation_nerfactor}, where it improves both relighting and albedo reconstruction.

\mysection{Material Regularization}
We measure the effect of the material regularization in \Table{ablation_nerfactor}.
By penalizing metalness prediction, our model tries to explain output radiance through roughness and albedo.
However, as visualized in \Figure{pulling}, the loss coefficient used is small enough to still allow our model to correctly predict metalness when required.

\mysection{Occlusion Averaging}
The occlusion factor we derive consists of a per-channel factor that depends on estimated lighting.
However, since both are being learned jointly, we observe that this can be a noisy process.
We find in \Table{ablation_nerfactor} that relighting and albedo reconstruction both improve when we supervise the occlusion factors $\hat{o}_d$ and $\hat{o}_s$ with their per-channel averages instead.
Assuming that all channels of the occlusion factor are equal is equivalent to assuming that all color channels of the lighting are equal, which slightly reduces the noise during training and uses fewer parameters.

\mysection{MLP vs. Mipmap}
We compare the effects of using a mipmap representation for illumination against our proposed regularized MLP.
As reported in \Table{ablation_nerfactor}, the mipmap representation leads to slight improvements in both relighting and albedo reconstruction.
However, these improvements come at a cost of efficiency, since the average training time for our method is only $47$ min., which increases to $64$ min. with the mipmap representation - a $36\%$ increase in training time.
As visualized in \Figure{envmaps}, our method produces smoother representations with less noise thanks to the MLP representation.
However, it is still capable of capturing fine details such as trees in the `materials' scene and buildings in the `teapot' scene.
We expect further improvements in the rapidly advancing field of neural rendering will translate to better MLP representations and benefit our method.


\section{Conclusion and Limitations}
\label{sec:Conclusion}

In conclusion, we present a novel and efficient method for inverse rendering based on neural surface rendering and the split sum approximation for image-based lighting.
Owing to our proposed integrated illumination MLP, we can jointly estimate geometry, lighting, and material properties in under one hour using a single NVIDIA A100 GPU.
Additionally, we propose a way of supervising an occlusion factor for diffuse and specular lighting such that self-occlusions are accounted for with the split sum approximation.
Altogether, our method is capable of producing high-quality estimates of geometry, lighting, and material properties as measured by rendering objects under unseen views and lighting conditions.

However, due to the highly complex problem that inverse rendering presents, there are some limitations to our method.
The major assumptions we rely on come from using image-based lighting and the split sum approximation.
Image-based lighting assumes that light sources are located infinitely far away from the scene, leading to errors when this assumption is violated.
While we have tackled the problem of missing self-occlusions within the split sum approximation, we ignore the lack of indirect illumination.
Additionally, we only consider the reflection of light and are unable to model transmission and subsurface scattering effects.
We hope future works will be able to tackle some of these limitations.

{
    \small
    \bibliographystyle{ieeenat_fullname}
    \bibliography{main}
}

\maketitlesupplementary
\section{Appendix}


\subsection{Derivation of Illumination Loss}
In this section, we go through the derivation for the Monte Carlo approximation of pre-integrated illumination $\bar{g}$ used in \Equation{gt_roughness}.
We first split the specular light integral into two terms:
\begin{equation}
    \mathbf{L}_s = \frac{\int\limits_\Omega \mathbf{f}_s \mathbf{L}_i \langle \mathbf{\omega}_i,  \mathbf{n} \rangle d\mathbf{\omega}_i}{\int\limits_\Omega \mathbf{f}_s \langle \mathbf{\omega}_i,  \mathbf{n} \rangle d\mathbf{\omega}_i} \int\limits_\Omega \mathbf{f}_s \langle \mathbf{\omega}_i, \mathbf{n} \rangle d\mathbf{\omega}_i .
\label{eq:original_split_sum}
\end{equation}

As mentioned in the main paper, the term on the right can be precomputed so we focus on calculating an approximation for the term on the left.
\begin{equation}
g(\mathbf{\omega}_r, \rho) \approx \frac{\int\limits_\Omega \mathbf{f}_s \mathbf{L}_i \langle \mathbf{\omega}_i,  \mathbf{n} \rangle d\mathbf{\omega}_i}{\int\limits_\Omega \mathbf{f}_s \langle \mathbf{\omega}_i,  \mathbf{n} \rangle d\mathbf{\omega}_i}.
\label{eq:left_term}
\end{equation}

This term requires us to make two approximations to the Cook-Torrance BRDF $\mathbf{f}_s$

\begin{equation}
    \mathbf{f}_s = \frac{D F G}{4 \langle \mathbf{\omega_o}, \mathbf{n} \rangle \langle \mathbf{\omega_i}, \mathbf{n} \rangle } ,
\label{eq:brdf}
\end{equation}

\noindent to be able to approximate $g$ as blurred environment maps as per the split sum approximation.
The first approximation on the BRDF consists of assuming the multiplication between fresnel and geometric shadowing terms is approximately equal to the dot product between the normal and viewing directions: $FG \approx \langle \mathbf{\omega}_i, \mathbf{n} \rangle$.
Thus, we have that

\begin{equation}
    \mathbf{f}_s \approx \frac{D}{4 \langle \mathbf{\omega}_o, \mathbf{n} \rangle} , \quad
    g(\mathbf{\omega}_r, \rho) \approx \frac{\int\limits_\Omega D \mathbf{L}_i \langle \mathbf{\omega}_i,  \mathbf{n} \rangle d\mathbf{\omega}_i}{\int\limits_\Omega D \langle \mathbf{\omega}_i,  \mathbf{n} \rangle d\mathbf{\omega}_i},
\label{eq:brdf_approx_2}
\end{equation}

\noindent as shown in \Equation{split_sum}.
We use the GGX (Trowbridge-Reitz) microfacet distribution function for $D$: 

\begin{equation}
    D(\mathbf{\omega}_i, \mathbf{\omega}_o, \mathbf{n}, \rho) = \frac{\rho^2}{\pi (\langle \mathbf{h}, \mathbf{n} \rangle^2 (\rho^2 - 1) + 1)^2} ,
\label{eq:microfacet_distribution}
\end{equation}

\noindent where $\mathbf{h}$ is the half vector between $\mathbf{\omega}_i$ and  $\mathbf{\omega}_o$.
The second approximation assumes the normal and viewing directions to be equal to the reflection direction.
That is, $\mathbf{n} \approx \mathbf{\omega}_r$ and $\mathbf{\omega}_o \approx \mathbf{\omega}_r$.
This leaves us with the following simplified $D$:

\begin{equation}
    D(\mathbf{\omega}_i, \mathbf{\omega}_r, \rho) \approx \frac{\rho^2}{\pi (\frac{1 + \langle \mathbf{\omega}_i, \mathbf{\omega}_r \rangle}{2} (\rho^2 - 1) + 1)^2},
\label{eq:microfacet_distribution_approx}
\end{equation}


\noindent which now does not depend on either the normal or viewing directions.
Approximating both integrals with Monte Carlo sampling and taking the same number of samples, we arrive at the expression in \Equation{gt_roughness}:

\begin{equation}
    \bar{g}(\mathbf{\omega}_r, \rho) = \frac{\sum\limits_\Omega D(\mathbf{\omega}_i, \mathbf{\omega}_r, \rho) \mathbf{L}_i \langle \mathbf{\omega}_i,  \mathbf{\omega}_r \rangle d\mathbf{\omega}_i}{\sum\limits_\Omega D(\mathbf{\omega}_i, \mathbf{\omega}_r, \rho) \langle \mathbf{\omega}_i,  \mathbf{\omega}_r \rangle d\mathbf{\omega}_i}.
\label{eq:monte_carlo_illum}
\end{equation}

\subsection{Derivation of Occlusion Factor Approximation}
We now go over the derivation of the occlusion factor Monte Carlo approximation.
As shown in \Equation{diffuse_visibility_factor}, we aim to approximate the occlusion factors $\mathbf{o}_d(\mathbf{x})$ and $\mathbf{o}_s(\mathbf{x})$

\begin{equation}
    \mathbf{o}_d(\mathbf{x}) = \frac{\int\limits_\Omega  \mathbf{L}_i V_i \langle \mathbf{\omega}_i, \mathbf{n} \rangle d\mathbf{\omega}_i }{\int\limits_\Omega  \mathbf{L}_i \langle \mathbf{\omega}_i, \mathbf{n} \rangle d\mathbf{\omega}_i } ,
\label{eq:occlusion_factor_main}
\end{equation}

\begin{equation}
    \mathbf{o}_s(\mathbf{x}) = \frac{\int\limits_\Omega  D \mathbf{L}_i V_i \langle \mathbf{\omega}_i, \mathbf{n} \rangle d\mathbf{\omega}_i }{\int\limits_\Omega D  \mathbf{L}_i \langle \mathbf{\omega}_i, \mathbf{n} \rangle d\mathbf{\omega}_i } ,
\label{eq:occlusion_factor_specular}
\end{equation}

\noindent in order to supervise an occlusion factor network through Monte Carlo sampling.
As shown in the main manuscript, we approximate $o_d$ with Monte Carlo sampling by taking the same number of samples for both integrals:

\begin{equation}
    \bar{\mathbf{o}}_d(\mathbf{x}) = \frac{\sum\limits_{\mathbf{\omega}_i \in \Omega} \mathbf{L}_i V_i}{\sum\limits_{\mathbf{\omega}_i \in \Omega} \mathbf{L}_i},
\label{eq:diffuse_occlusion_factor_suppl}
\end{equation}

\noindent with $\mathbf{\omega}_i$ taken from a cos-weighted sampling of the hemisphere around the normal $\mathbf{n}$ at location $\mathbf{x}$.
The probability density function sampled is given by:

\begin{equation}
\text{pdf}(\mathbf{\omega}_i) = \frac{\text{cos}(\theta)}{\pi},
\end{equation}

\noindent where $\theta$ is the angle between $\mathbf{\omega}_i$ and the normal. 
This cos-weighted sampling aids in reducing variance by eliminating the dot product factor from the estimation.
A similar derivation can be followed for the specular occlusion term leading to the following Monte Carlo estimate $\bar{\mathbf{o}}_s(\mathbf{x})$:

\begin{equation}
    \bar{\mathbf{o}}_s(\mathbf{x}) = \frac{\sum\limits_{\mathbf{\omega}_i \in \Omega} \mathbf{L}_i V_i \langle \mathbf{\omega}_i, \mathbf{n} \rangle }{\sum\limits_{\mathbf{\omega}_i \in \Omega} \mathbf{L}_i \langle \mathbf{\omega}_i, \mathbf{n} \rangle},
\label{eq:specular_occlusion_factor_suppl}
\end{equation}

\noindent where $\mathbf{\omega}_i$ is now obtained by sampling the GGX distribution in order to reduce variance by eliminating the factor $D$ from both integrals.
The probability density function sampled in this case is the following:


\begin{equation}
\text{pdf}(\mathbf{\omega}_i) = \frac{D (\mathbf{\omega}_i, \mathbf{\omega}_r, \rho) }{4 },
\end{equation}

\noindent which relies on the second approximation used in the previous section.

\subsection{Network Implementation Details}
We implement the spatial network using the progressive hash grid encoding from ~\cite{li2023neuralangelo}.
The hash grid consists of $16$ levels with $2$ features per level and a hashmap size of $2^{19}$ entries.
The base grid spatial resolution is $32$ voxels, increasing by ${\sim}1.32$ each level.
An MLP with a single $64$-channel hidden layer is used to produce spatial features with $13$ channels along with the SDF predictions.
Spatial features are then input to an MLP with two hidden layers of $256$ channels each and ReLU activations to produce material property (metalness, roughness, and albedo) predictions.
A separate but identical MLP is used to produce occlusion factor predictions.
A sigmoid is used to map the MLP outputs to the occlusion factor and material properties' ranges of $[0, 1]$.
The illumination network consists of an MLP with five hidden layers with $256$ channels each and ReLU activations.
Both the direction and roughness vectors used as input to the illumination network are first positionally encoded as proposed in ~\cite{mildenhall2021nerf}, using $10$ frequencies for the directional input and $5$ for the roughness input.
A softplus function is used to map the illumination network's output to the range $(0,\inf)$.

\subsection{Blender and Shiny Blender Relighting Results}
\begin{figure}[htb]
 \centering
 \includegraphics[width=1.\linewidth,trim={0cm 0cm 0cm 0cm},clip]{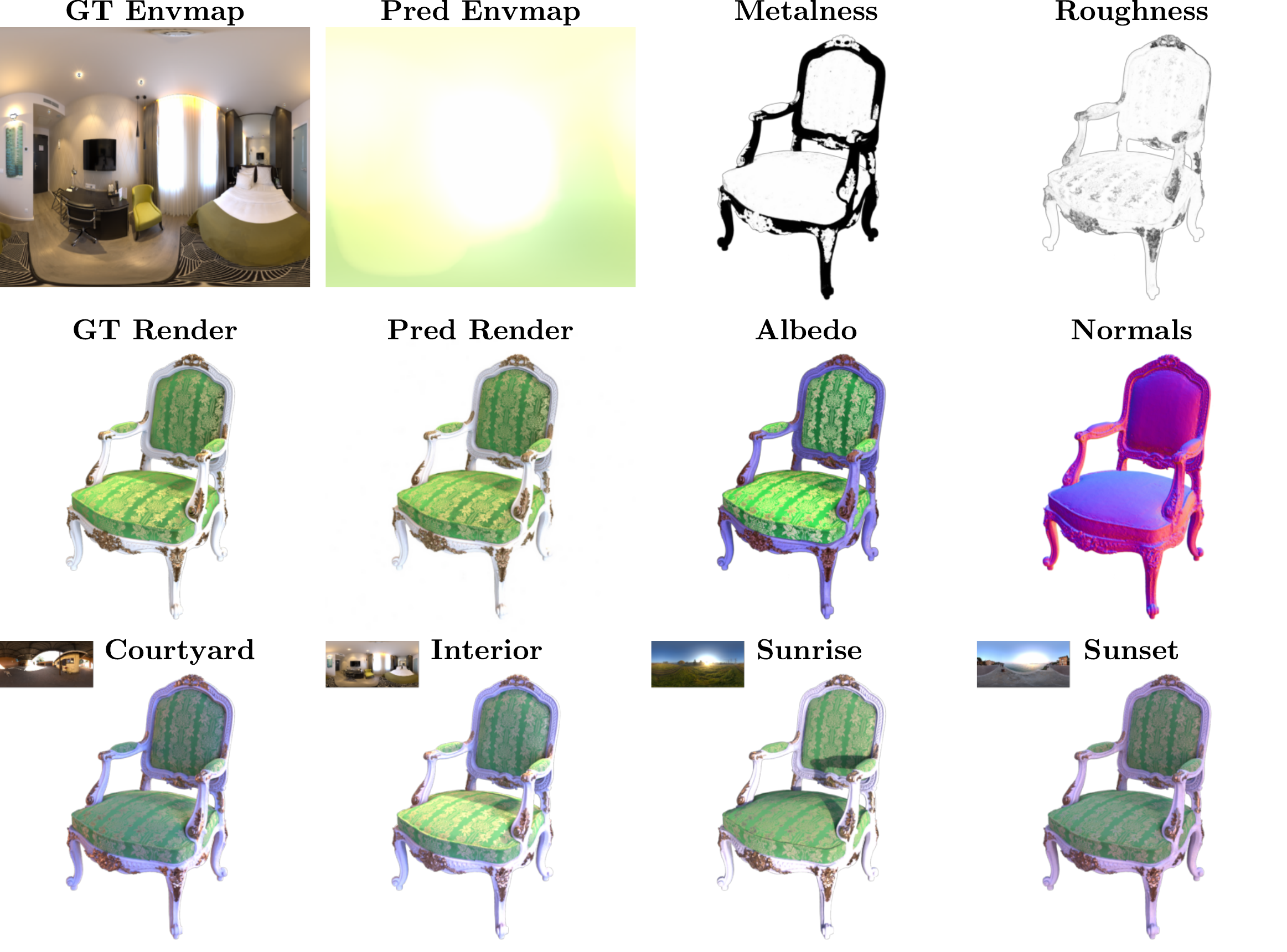}
 \caption{
     \textbf{Qualitative results on the Blender `chair' scene.} 
 }
 \label{fig:qual_chair}
\end{figure}

\begin{figure}[htb]
 \centering
 \includegraphics[width=1.\linewidth,trim={0cm 0cm 0cm 0cm},clip]{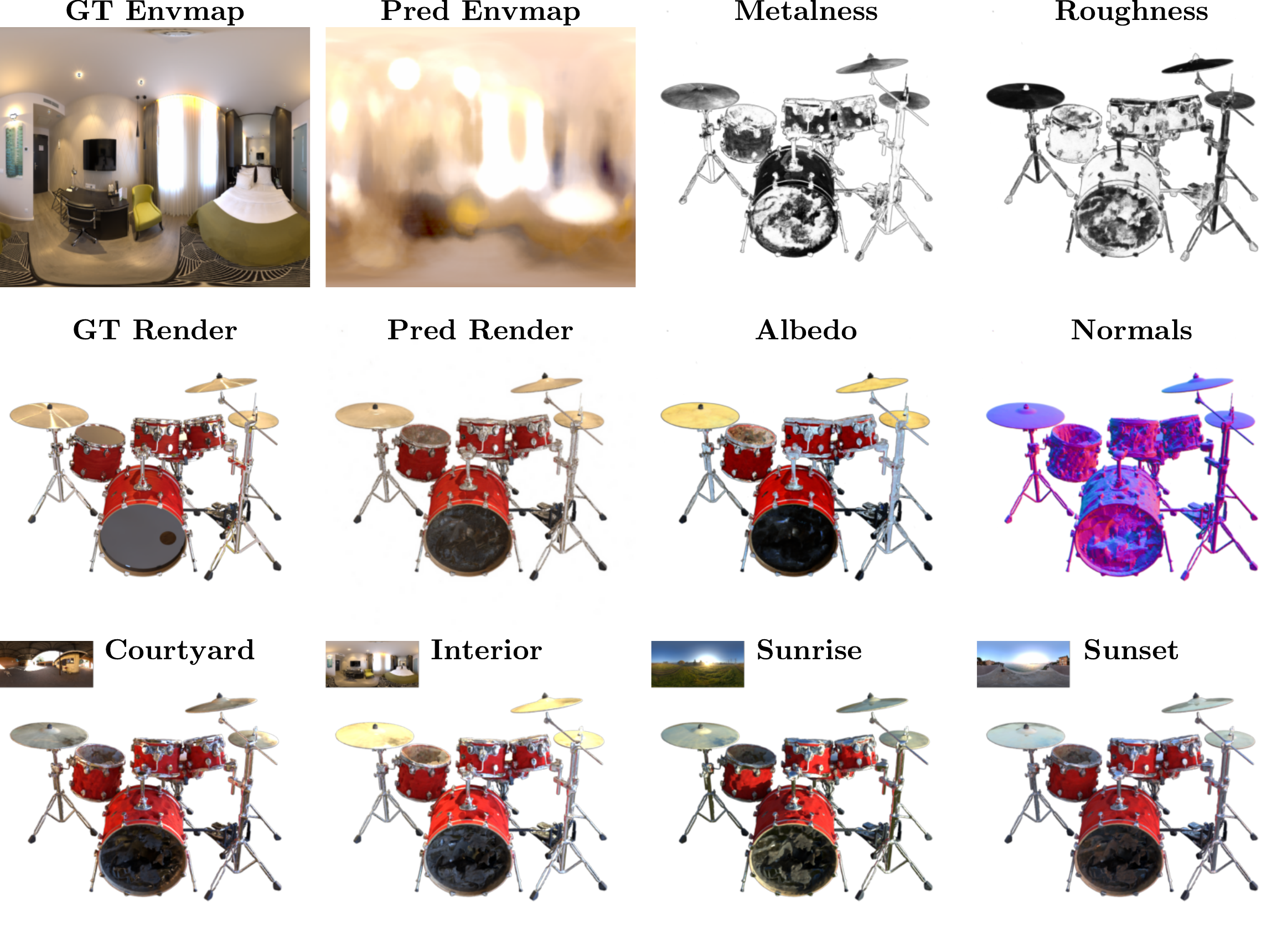}
 \caption{
     \textbf{Qualitative results on the Blender `drums' scene.} 
 }
 \label{fig:qual_drums}
\end{figure}

\begin{figure}[htb]
 \centering
 \includegraphics[width=1.\linewidth,trim={0cm 0cm 0cm 0cm},clip]{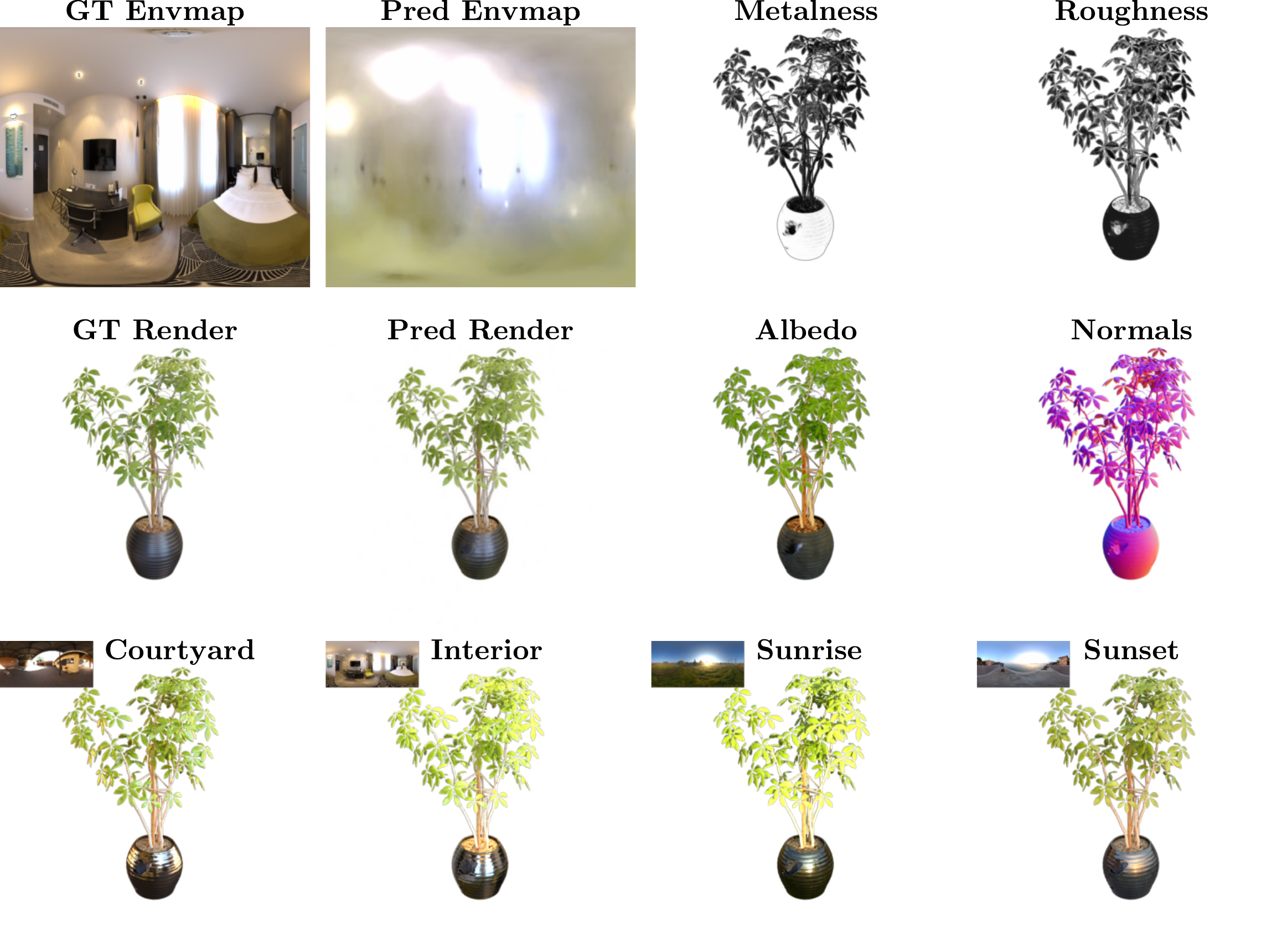}
 \caption{
     \textbf{Qualitative results on the Blender `ficus' scene.} 
 }
 \label{fig:qual_ficus}
\end{figure}

\begin{figure}[htb]
 \centering
 \includegraphics[width=1.\linewidth,trim={0cm 0cm 0cm 0cm},clip]{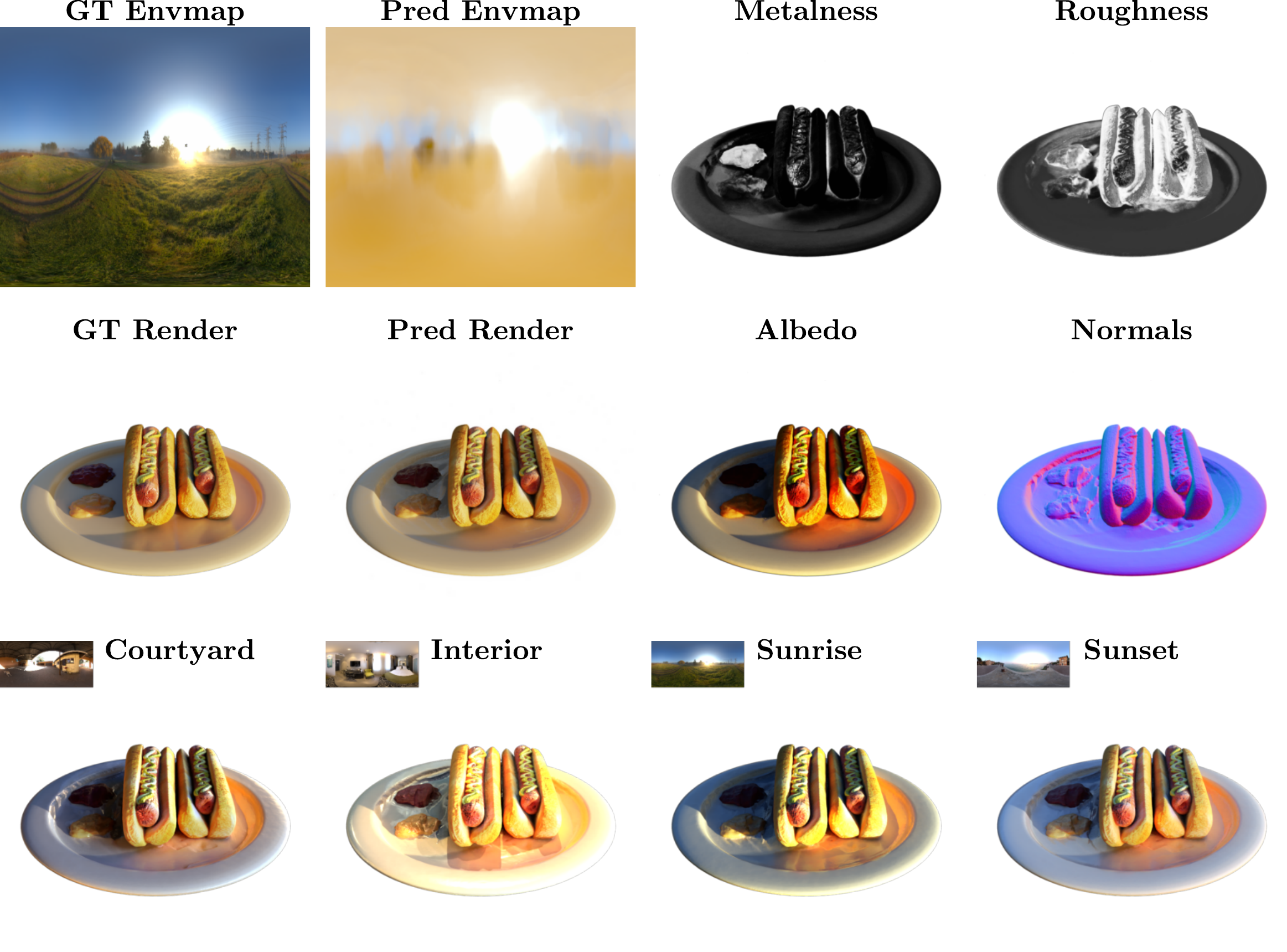}
 \caption{
     \textbf{Qualitative results on the Blender `hotdog' scene.} 
 }
 \label{fig:qual_hotdog}
\end{figure}

\begin{figure}[htb]
 \centering
 \includegraphics[width=1.\linewidth,trim={0cm 0cm 0cm 0cm},clip]{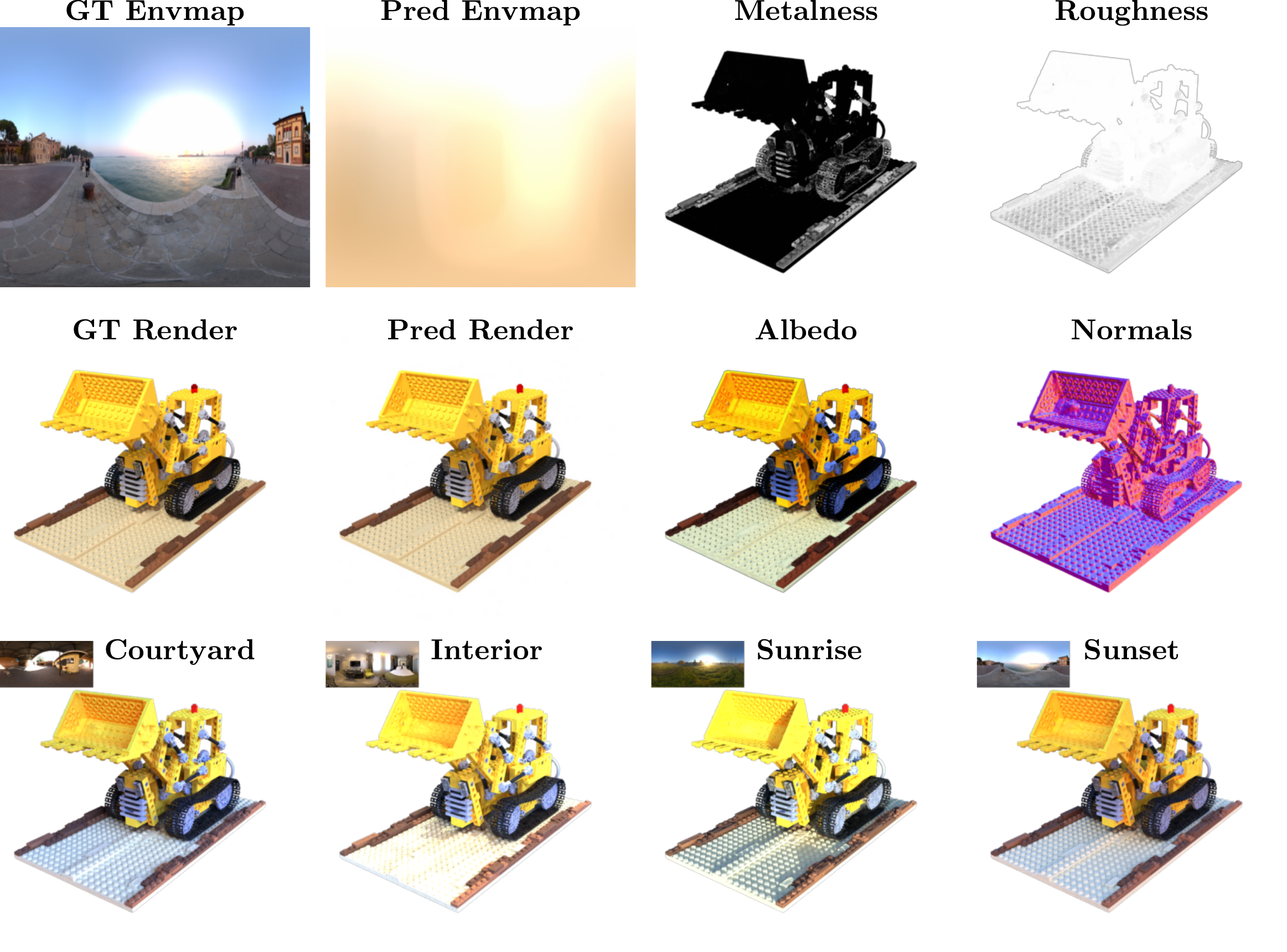}
 \caption{
     \textbf{Qualitative results on the Blender `lego' scene.} 
 }
 \label{fig:qual_lego}
\end{figure}

\begin{figure}[htb]
 \centering
 \includegraphics[width=1.\linewidth,trim={0cm 0cm 0cm 0cm},clip]{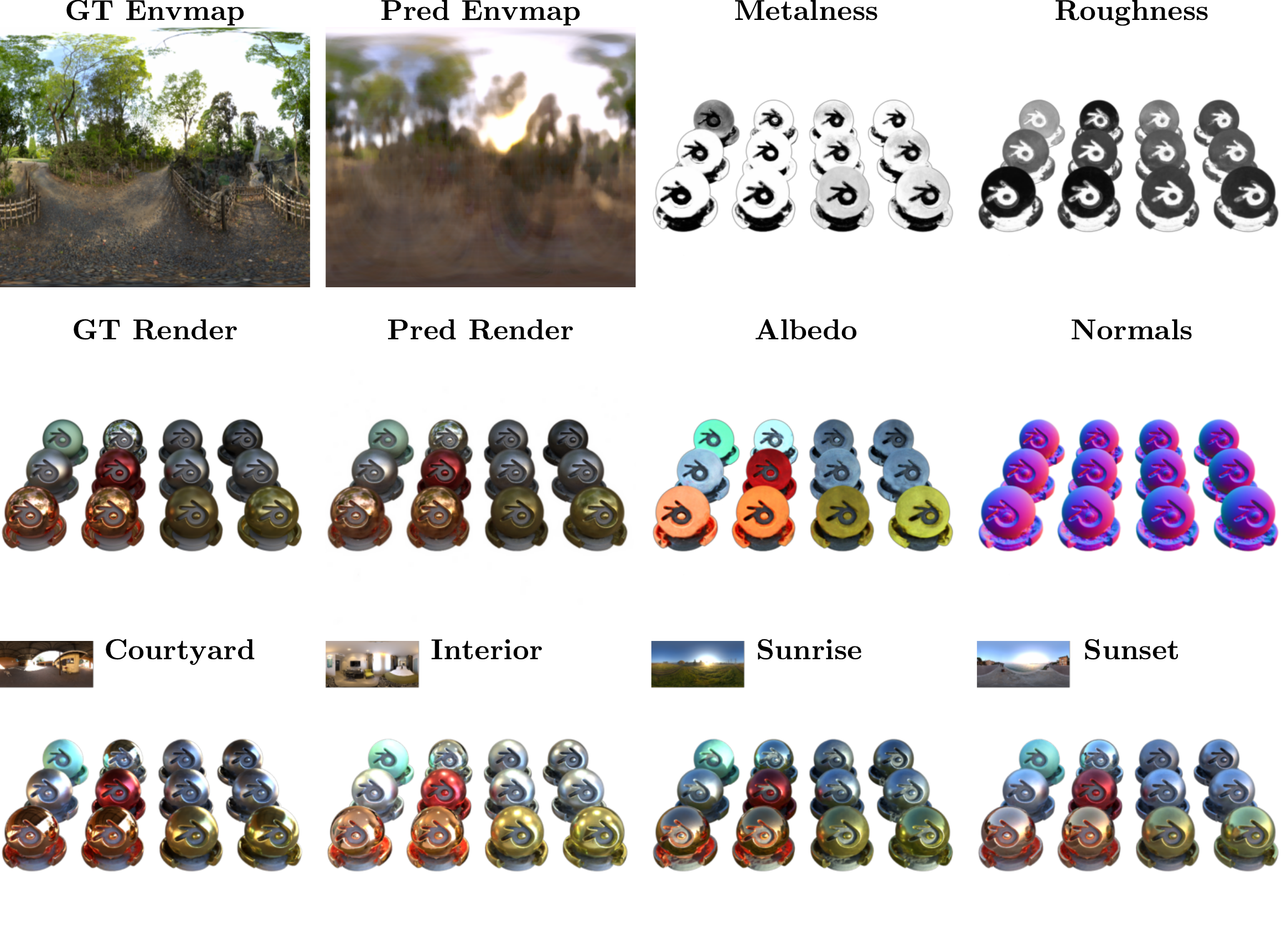}
 \caption{
     \textbf{Qualitative results on the Blender `materials' scene.} 
 }
 \label{fig:qual_materials}
\end{figure}

\begin{figure}[htb]
 \centering
 \includegraphics[width=1.\linewidth,trim={0cm 0cm 0cm 0cm},clip]{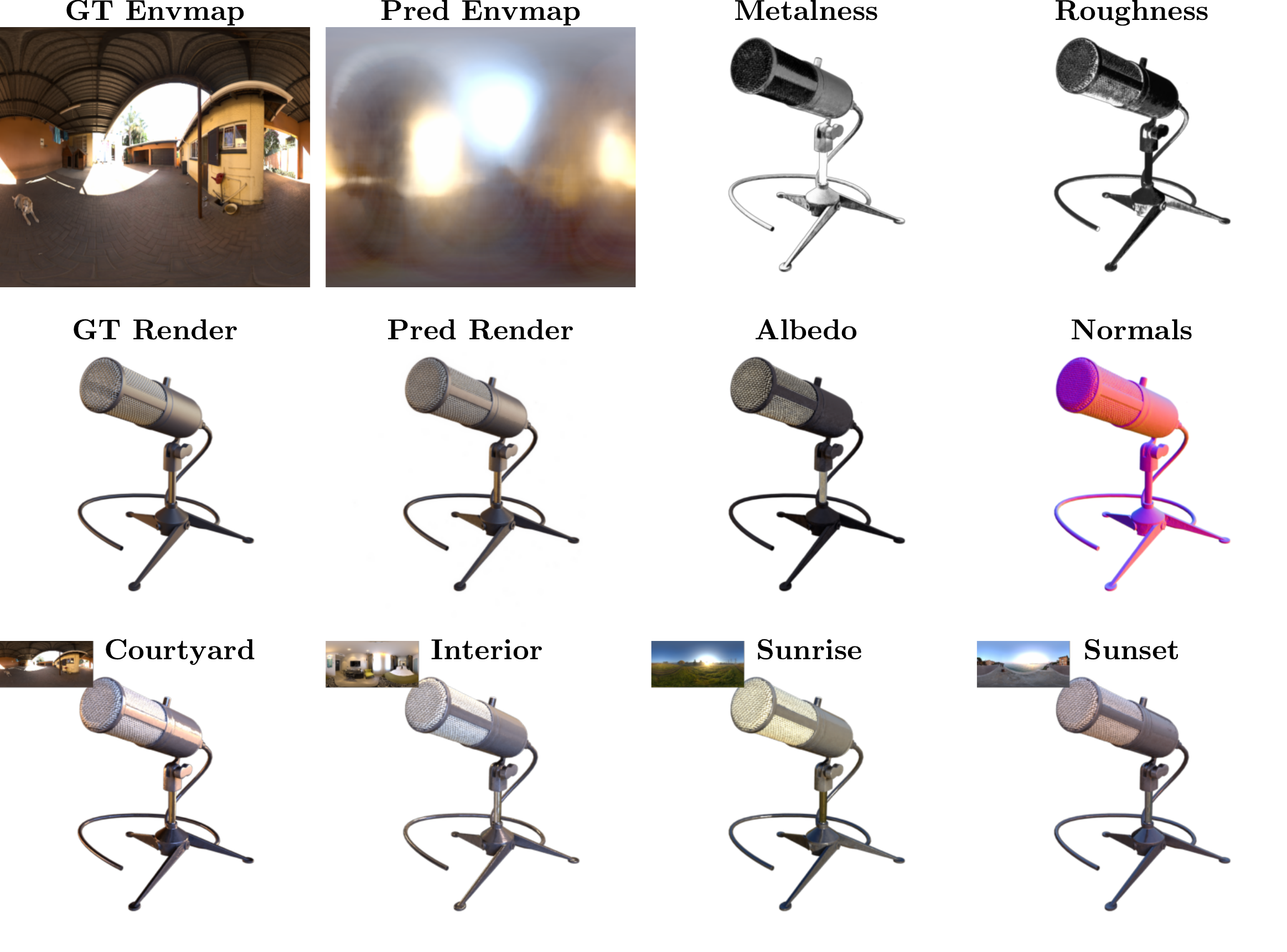}
 \caption{
     \textbf{Qualitative results on the Blender `mic' scene.} 
 }
 \label{fig:qual_mic}
\end{figure}

\begin{figure}[htb]
 \centering
 \includegraphics[width=1.\linewidth,trim={0cm 0cm 0cm 0cm},clip]{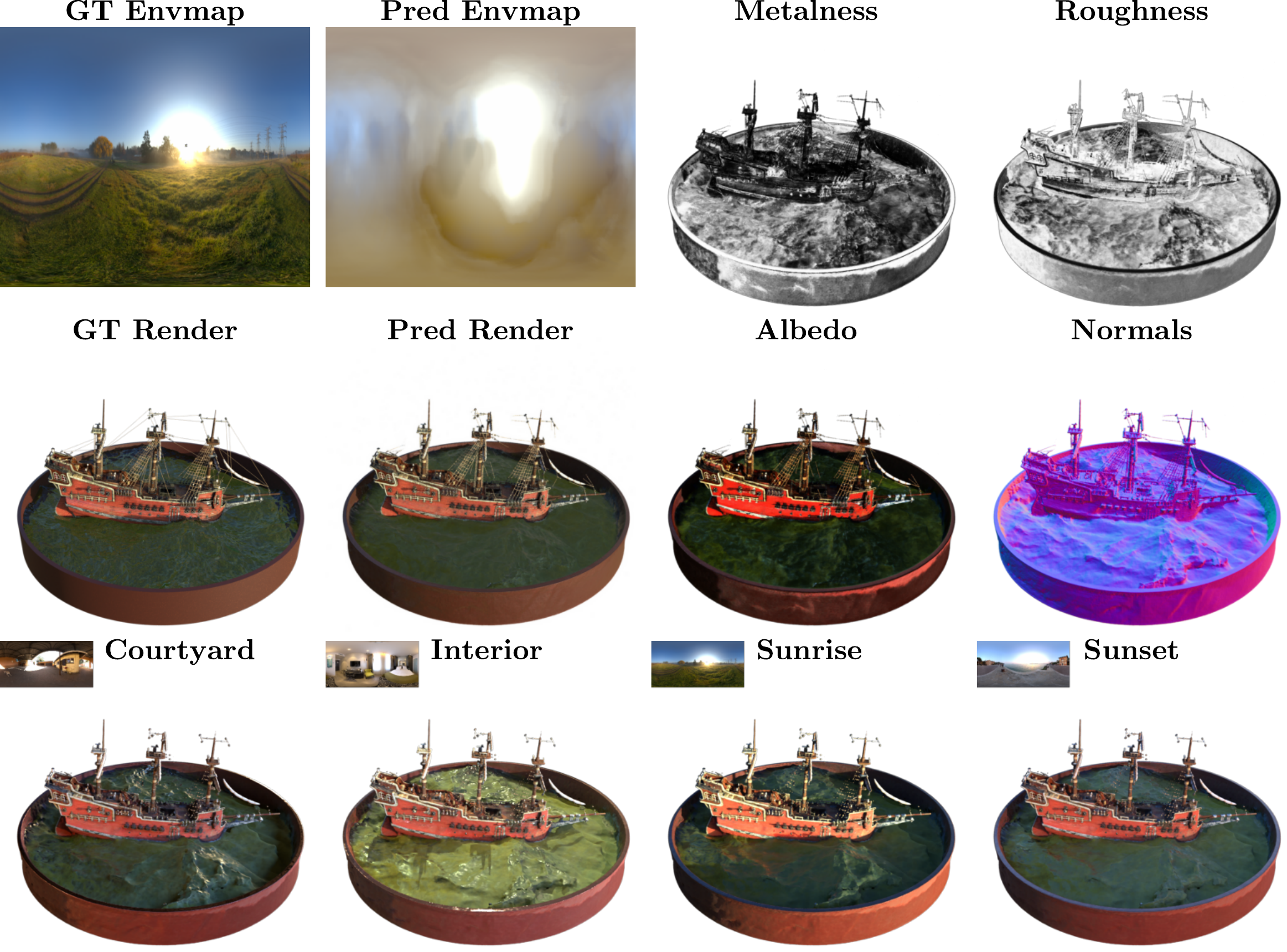}
 \caption{
     \textbf{Qualitative results on the Blender `ship' scene.} 
 }
 \label{fig:qual_ship}
\end{figure}

\begin{figure}[htb]
 \centering
 \includegraphics[width=1.\linewidth,trim={0cm 0cm 0cm 0cm},clip]{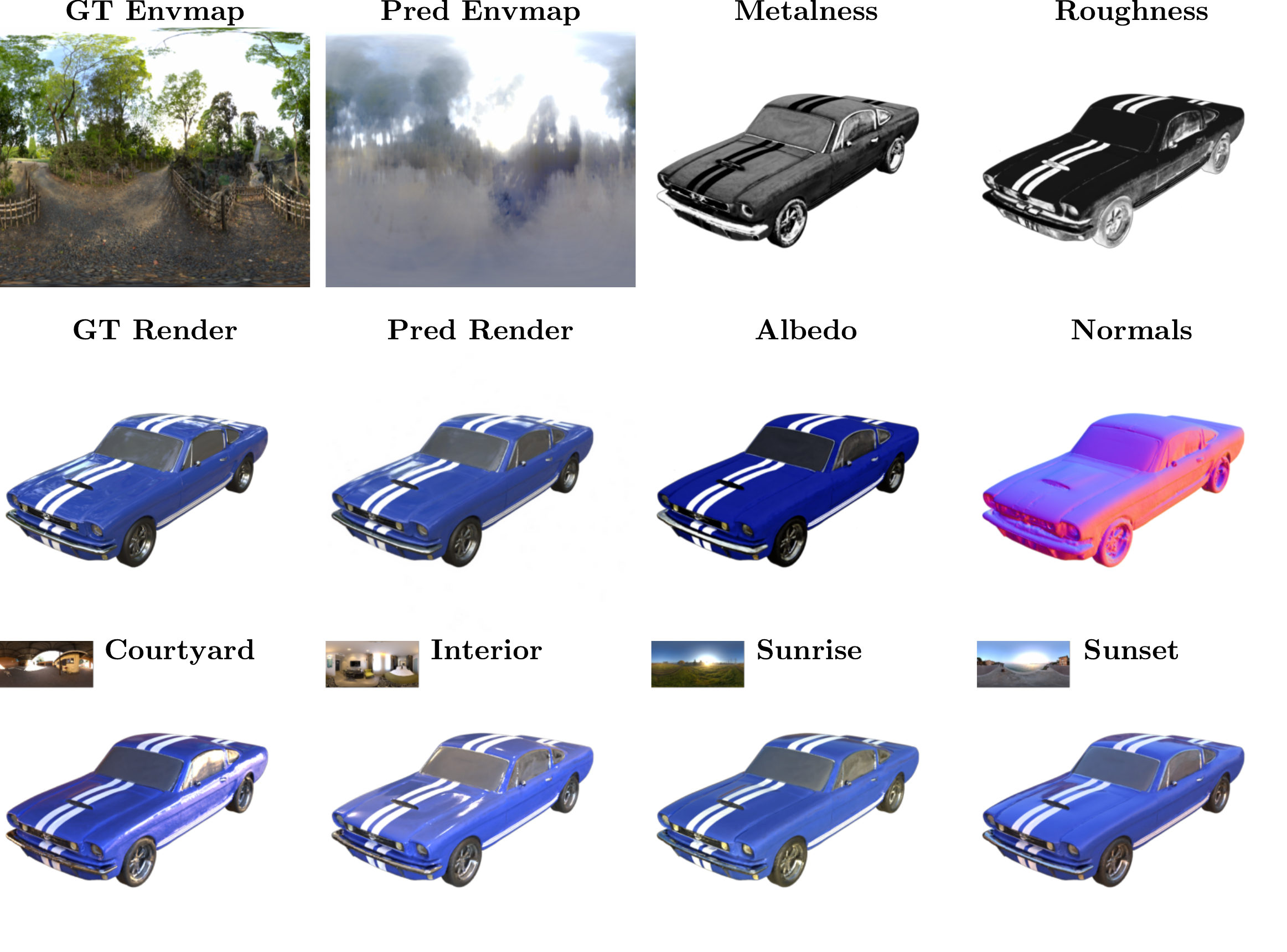}
 \caption{
     \textbf{Qualitative results on the Shiny Blender `car' scene.} 
 }
  \label{fig:qual_car}
\end{figure}

\begin{figure}[htb]
 \centering
 \includegraphics[width=1.\linewidth,trim={0cm 0cm 0cm 0cm},clip]{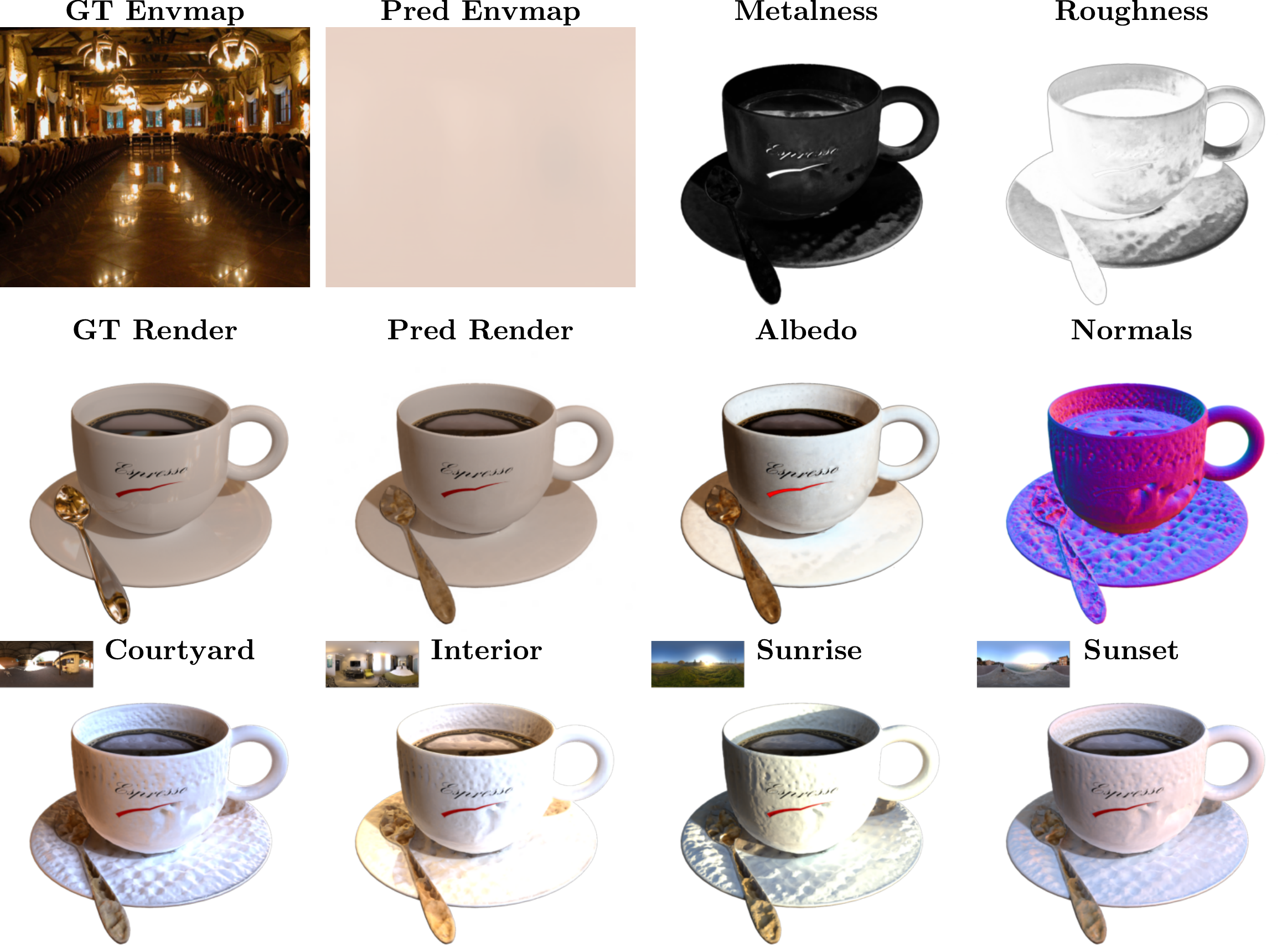}
 \caption{
     \textbf{Qualitative results on the Shiny Blender `coffee' scene.} 
 }
  \label{fig:qual_coffee}
\end{figure}

\begin{figure}[htb]
 \centering
 \includegraphics[width=1.\linewidth,trim={0cm 0cm 0cm 0cm},clip]{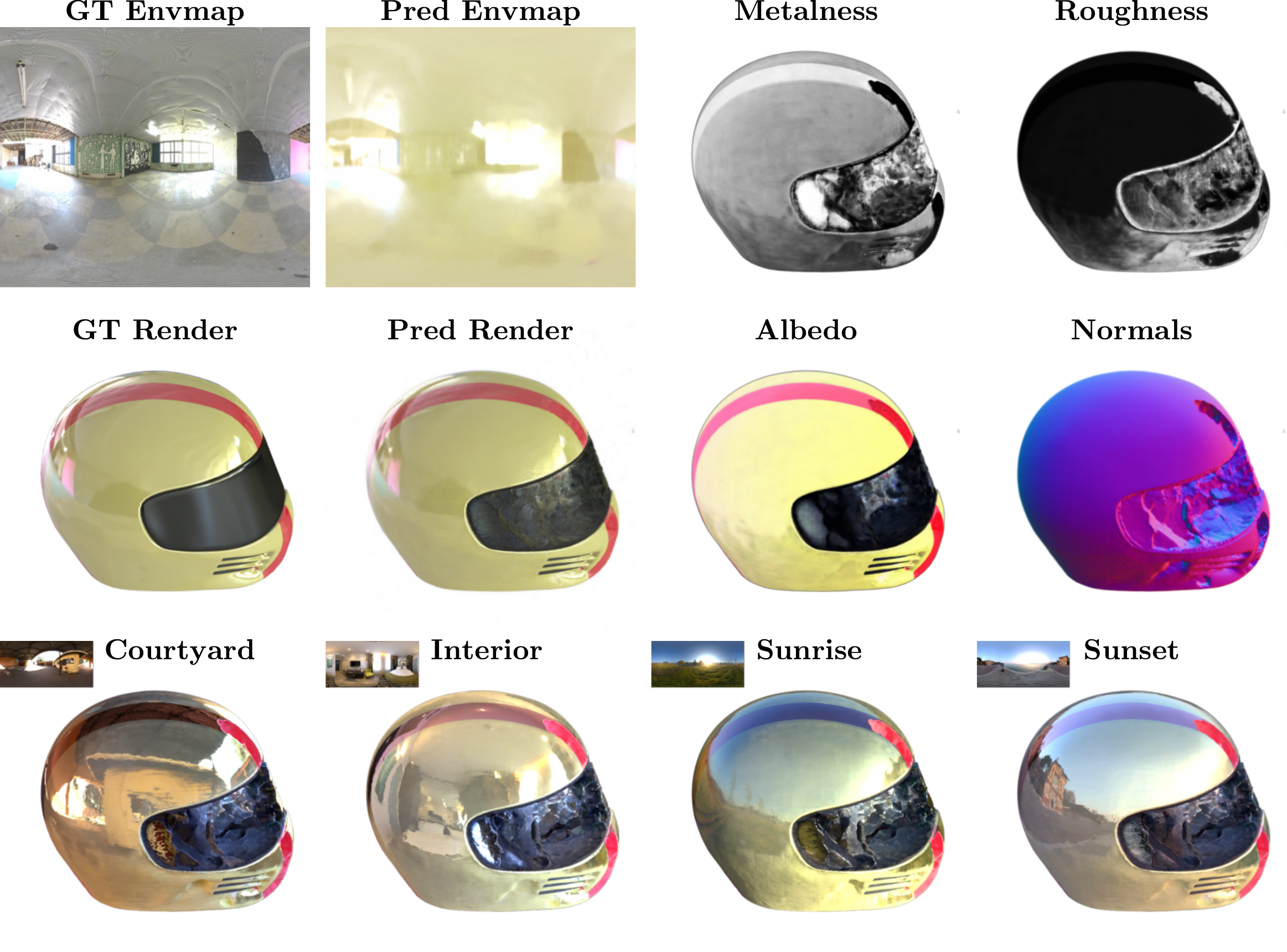}
 \caption{
     \textbf{Qualitative results on the Shiny Blender `helmet' scene.} 
 }
  \label{fig:qual_helmet}
\end{figure}

\begin{figure}[htb]
 \centering
 \includegraphics[width=1.\linewidth,trim={0cm 0cm 0cm 0cm},clip]{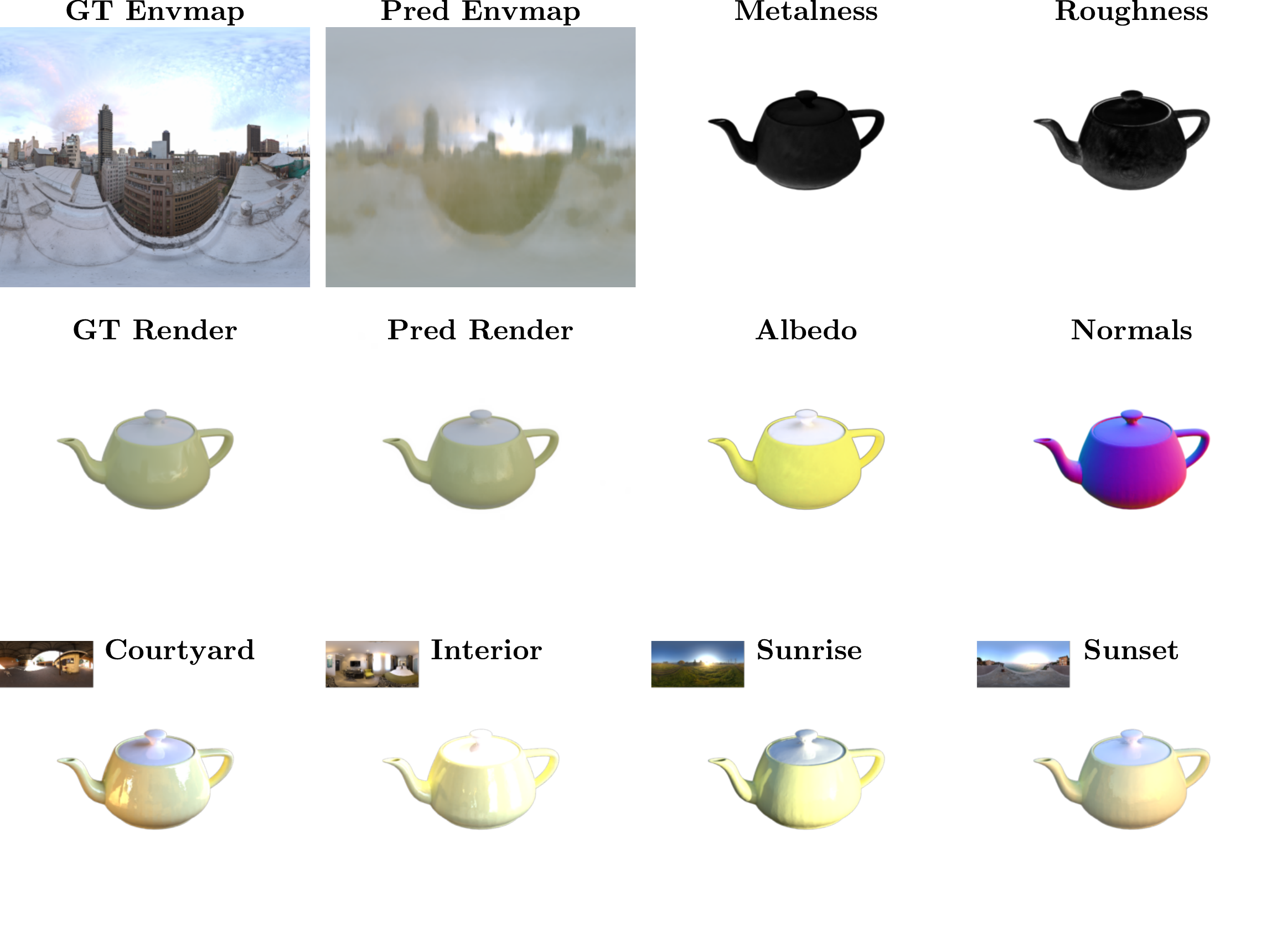}
 \caption{
     \textbf{Qualitative results on the Shiny Blender `teapot' scene.} 
 }
  \label{fig:qual_teapot}
\end{figure}

\begin{figure}[htb]
 \centering
 \includegraphics[width=1.\linewidth,trim={0cm 0cm 0cm 0cm},clip]{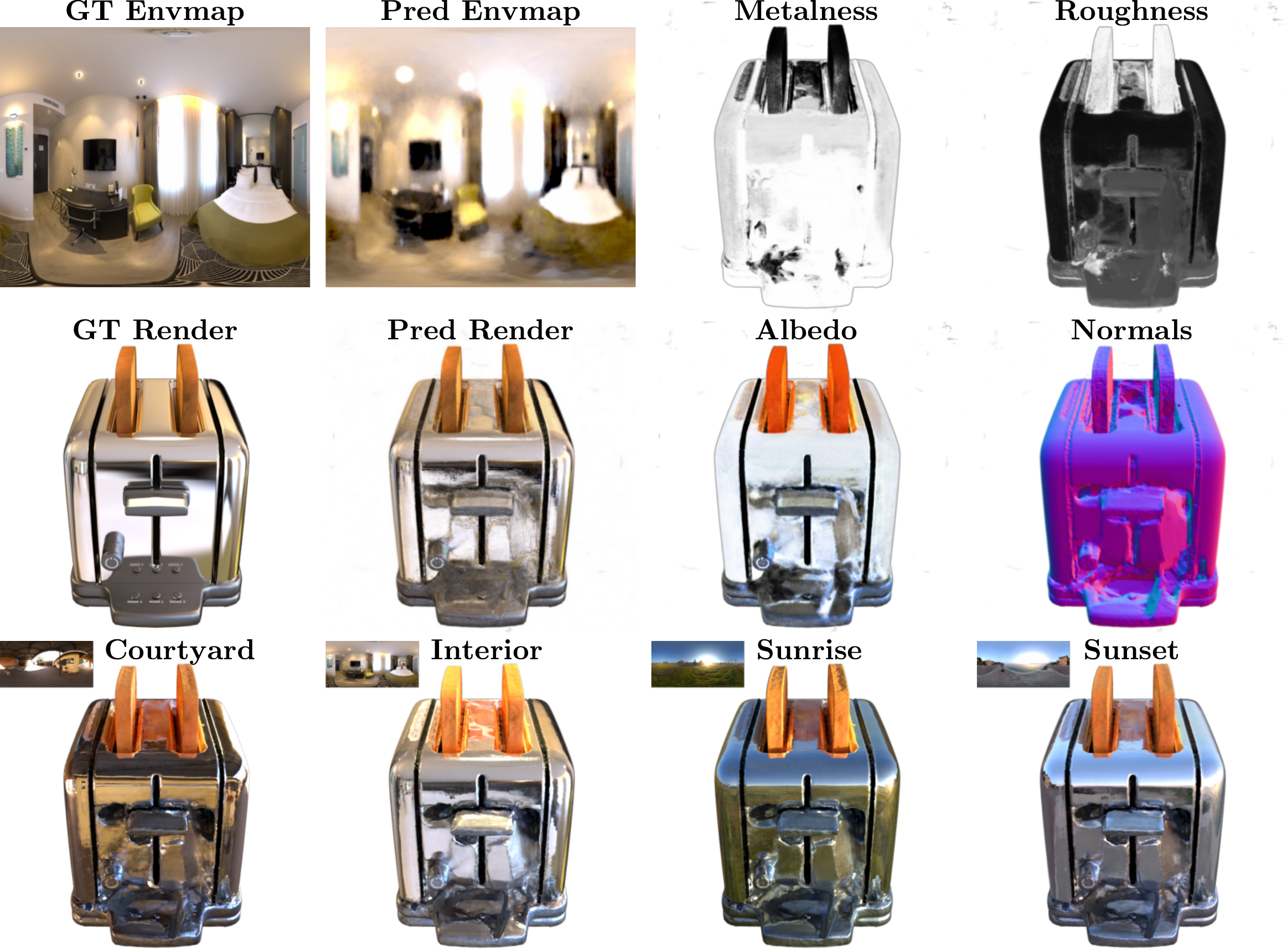}
 \caption{
     \textbf{Qualitative results on the Shiny Blender `toaster' scene.} 
 }
  \label{fig:qual_toaster}
\end{figure}

We report per-scene metrics for relighting using the Blender dataset in \Cref{tab:blender_relighting_psnr,tab:blender_relighting_SSIM,tab:blender_relighting_LPIPS}, and using the Shiny Blender dataset in \Cref{tab:shiny_blender_relighting_psnr,tab:shiny_blender_relighting_SSIM,tab:shiny_blender_relighting_LPIPS}.
Additionally, we present qualitative results of our method visualizing the learnt illumination, material properties (metalness, roughness, and albedo), geometry, and relit renderings from our method's predictions for the Blender dataset in \Cref{fig:qual_chair,fig:qual_drums,fig:qual_ficus,fig:qual_hotdog,fig:qual_lego,fig:qual_materials,fig:qual_mic,fig:qual_ship} and for the Shiny Blender dataset in \Cref{fig:qual_car,fig:qual_coffee,fig:qual_helmet,fig:qual_teapot,fig:qual_toaster}.

\begin{table*}[htb]
   \centering{
        \csvreader[
            tabular=l|c|cccccccc,
            table head={
                \Xhline{2\arrayrulewidth}
                & \multicolumn{9}{c}{$\mathbf{PSNR \uparrow}$} \\
                \Xhline{2\arrayrulewidth}
                \bfseries & \bfseries avg. & \bfseries chair & \bfseries drums & \bfseries ficus & \bfseries hotdog & \bfseries lego & \bfseries materials & \bfseries mic & \bfseries ship
                \\\Xhline{2\arrayrulewidth}
            },
            table foot=\Xhline{2\arrayrulewidth},
            head to column names,
        ]{results/relighting_nerf.csv}{}
        {\bfseries\Name & \avgPSNRNerf & \chairPSNRNerf & \drumsPSNRNerf & \ficusPSNRNerf & \hotdogPSNRNerf & \legoPSNRNerf & \materialsPSNRNerf & \micPSNRNerf & \shipPSNRNerf} 
    }
    \caption{\textbf{Blender per-scene PSNR.} }
    \label{tab:blender_relighting_psnr}
\end{table*}

\begin{table*}[htb]
    \centering{
        \csvreader[
            tabular=l|c|cccccccc,
            table head={
                \Xhline{2\arrayrulewidth}
                & \multicolumn{9}{c}{$\mathbf{SSIM \uparrow}$} \\
                \Xhline{2\arrayrulewidth}
                \bfseries & \bfseries avg. & \bfseries chair & \bfseries drums & \bfseries ficus & \bfseries hotdog & \bfseries lego & \bfseries materials & \bfseries mic & \bfseries ship
                \\\Xhline{2\arrayrulewidth}
            },
            table foot=\Xhline{2\arrayrulewidth},
            head to column names,
        ]{results/relighting_nerf.csv}{}
        {\bfseries\Name & \avgSSIMNerf & \chairSSIMNerf & \drumsSSIMNerf & \ficusSSIMNerf & \hotdogSSIMNerf & \legoSSIMNerf & \materialsSSIMNerf & \micSSIMNerf & \shipSSIMNerf} 
    }
    \caption{\textbf{Blender per-scene SSIM.} }
    \label{tab:blender_relighting_SSIM}
\end{table*}

\begin{table*}[htb]
    \centering{
        \csvreader[
            tabular=l|c|cccccccc,
            table head={
                \Xhline{2\arrayrulewidth}
                & \multicolumn{9}{c}{$\mathbf{LPIPS \downarrow}$} \\
                \Xhline{2\arrayrulewidth}
                \bfseries & \bfseries avg. & \bfseries chair & \bfseries drums & \bfseries ficus & \bfseries hotdog & \bfseries lego & \bfseries materials & \bfseries mic & \bfseries ship
                \\\Xhline{2\arrayrulewidth}
            },
            table foot=\Xhline{2\arrayrulewidth},
            head to column names,
        ]{results/relighting_nerf.csv}{}
        {\bfseries\Name & \avgLPIPSNerf & \chairLPIPSNerf & \drumsLPIPSNerf & \ficusLPIPSNerf & \hotdogLPIPSNerf & \legoLPIPSNerf & \materialsLPIPSNerf & \micLPIPSNerf & \shipLPIPSNerf} 
    }
    \caption{\textbf{Blender per-scene LPIPS.} }
    \label{tab:blender_relighting_LPIPS}
\end{table*}

\begin{table*}[htb]
    \centering{
        \csvreader[
            tabular=l|c|ccccc,
            table head={
                \Xhline{2\arrayrulewidth}
                & \multicolumn{6}{c}{$\mathbf{PSNR \uparrow}$} \\
                \Xhline{2\arrayrulewidth}
                \bfseries & \bfseries avg. & \bfseries car & \bfseries coffee & \bfseries helmet & \bfseries teapot & \bfseries toaster
                \\\Xhline{2\arrayrulewidth}
            },
            table foot=\Xhline{2\arrayrulewidth},
            head to column names,
        ]{results/relighting_nerf.csv}{}
        {\bfseries\Name & \avgPSNRShiny & \carPSNRShiny & \coffeePSNRShiny & \helmetPSNRShiny & \teapotPSNRShiny & \toasterPSNRShiny} 
    }
    \caption{\textbf{Shiny Blender per-scene PSNR.} }
    \label{tab:shiny_blender_relighting_psnr}
\end{table*}

\begin{table*}[htb]
    \centering{
        \csvreader[
            tabular=l|c|ccccc,
            table head={
                \Xhline{2\arrayrulewidth}
                & \multicolumn{6}{c}{$\mathbf{SSIM \uparrow}$} \\
                \Xhline{2\arrayrulewidth}
                \bfseries & \bfseries avg. & \bfseries car & \bfseries coffee & \bfseries helmet & \bfseries teapot & \bfseries toaster
                \\\Xhline{2\arrayrulewidth}
            },
            table foot=\Xhline{2\arrayrulewidth},
            head to column names,
        ]{results/relighting_nerf.csv}{}
        {\bfseries\Name & \avgSSIMShiny & \carSSIMShiny & \coffeeSSIMShiny & \helmetSSIMShiny & \teapotSSIMShiny & \toasterSSIMShiny} 
    }
    \caption{\textbf{Shiny Blender per-scene SSIM.} }
    \label{tab:shiny_blender_relighting_SSIM}
\end{table*}

\begin{table*}[htb]
    \centering{
        \csvreader[
            tabular=l|c|ccccc,
            table head={
                \Xhline{2\arrayrulewidth}
                & \multicolumn{6}{c}{$\mathbf{LPIPS \downarrow}$} \\
                \Xhline{2\arrayrulewidth}
                \bfseries & \bfseries avg. & \bfseries car & \bfseries coffee & \bfseries helmet & \bfseries teapot & \bfseries toaster
                \\\Xhline{2\arrayrulewidth}
            },
            table foot=\Xhline{2\arrayrulewidth},
            head to column names,
        ]{results/relighting_nerf.csv}{}
        {\bfseries\Name & \avgLPIPSShiny & \carLPIPSShiny & \coffeeLPIPSShiny & \helmetLPIPSShiny & \teapotLPIPSShiny & \toasterLPIPSShiny} 
    }
    \caption{\textbf{Shiny Blender per-scene LPIPS.} }
    \label{tab:shiny_blender_relighting_LPIPS}
\end{table*}

\end{document}